\documentclass[10pt,twocolumn,letterpaper]{article}
\pdftrailerid{}

\usepackage{wacv}

\usepackage[normalem]{ulem}
\usepackage[accsupp]{axessibility}

\usepackage{siunitx}
\usepackage{colortbl}
\usepackage{enumitem}
\usepackage{multirow}
\usepackage[ruled,vlined,linesnumbered]{algorithm2e}

\definecolor{wacvblue}{rgb}{0.21,0.49,0.74}
\ifdefined\SupplementOnly
\usepackage[draft,breaklinks,colorlinks,allcolors=wacvblue]{hyperref}
\else
\usepackage[breaklinks,colorlinks,allcolors=wacvblue]{hyperref}
\fi
\hypersetup{%
  pdftitle={StreamClarify: Proactive Clarification for Grounded Streaming Video Question Answering},
  pdfauthor={Zhuchen Shao, Bo Chen, Klara Nahrstedt}%
}
\def\wacvPaperID{1249}
\def\confName{WACV}
\def\confYear{2027}

\providecommand{\eg}{e.g.\xspace}

\ifdefined\SupplementOnly
\title{Supplementary Material for\\ StreamClarify: Proactive Clarification for\\ Grounded Streaming Video Question Answering}
\else
\title{StreamClarify: Proactive Clarification for\\ Grounded Streaming Video Question Answering}
\fi

\author{%
{Zhuchen Shao}\quad{Bo Chen}\quad{Klara Nahrstedt}\\
{University of Illinois Urbana-Champaign}\\
{\tt\small{\{zhuchens,boc2,klara\}@illinois.edu}}\\
{\small{Project page: }\href{https://zhuchens-uiuc.github.io/StreamClarify-page/}{{\nolinkurl{zhuchens-uiuc.github.io/StreamClarify-page/}}}}%
}

\begin{document}

\AddToShipoutPicture*{%
  \AtPageLowerLeft{%
    \put(\LenToUnit{57bp},\LenToUnit{24bp}){%
      \parbox[b]{498bp}{\fontsize{7}{8}\selectfont
        \textcopyright~2027 IEEE. Personal use of this material is permitted.
        Permission from IEEE must be obtained for all other uses, in any current
        or future media, including reprinting/republishing this material for
        advertising or promotional purposes, creating new collective works,
        for resale or redistribution to servers or lists, or reuse of any
        copyrighted component of this work in other works.}%
    }%
  }%
}

\ifdefined\SupplementOnly
\twocolumn[{%
\centering
{\Large\bfseries Supplementary Material\par}
\vspace{0.35em}
{\large\bfseries StreamClarify: Proactive Clarification for Grounded Streaming Video Question Answering\par}
\vspace{0.9em}
}]
\else
\maketitle
\fi

\ifdefined\SupplementOnly
\raggedbottom
\appendix
\captionsetup{hypcap=false}
\setcounter{table}{6}
\setcounter{figure}{5}
\else

\begin{abstract}
Streaming video question answering (QA) requires real-time reasoning over continuous video streams under strict latency and active-context budgets. Current streaming vision-language models (VLMs) often give confident yet ungrounded answers when queries are ambiguous or have false premises, or when relevant visual evidence is no longer in the active context. Proactive clarification can help, but must satisfy the same latency and memory constraints. We present StreamClarify, a training-free framework for grounded streaming QA that combines a pre-answer uncertainty check with hierarchical memory. The check uses the entropy of a binary pre-answer judgment as a lightweight uncertainty score to decide whether to answer directly or request clarification. To support re-answering, a temporal-pyramid KV cache maintains a bounded active visual context, while a CPU-side semantic retrieval archive provides access to historical frames. We further construct StreamClarify-QA, a diagnostic benchmark of 1,837 QA pairs from StreamBench and OVO-Bench, spanning standard, referential ambiguity, and unanswerable scenarios. Across three VLM backbones, StreamClarify improves Macro accuracy by up to 8.2 points over backbone-matched baselines. At comparable trigger rates, entropy-based selection achieves more than 18 points higher precision than random selection in identifying initially incorrect answers. Under oracle user guidance, clarification improves Macro accuracy by up to 3.5 points. We also report a {}{controlled, oracle-assisted} Qwen3-VL simulation on Ambiguous QA with generated questions and referent-only user replies, where the entropy-gated condition improves accuracy by 8.3 points.
\end{abstract}

\section{Introduction}

Streaming video question answering (QA) places vision-language models (VLMs)
in an always-on assistant setting, requiring real-time reasoning over continuous,
unsegmented video streams~\cite{chen2024videollm,xu2025streamingvlm}. Unlike
offline video QA on pre-segmented clips~\cite{fu2025video,wang2025lvbench,
wu2024longvideobench}, streaming assistants process frames in a single online
pass within a bounded active-context
{}{budget. Recent work focuses on memory-efficient}
long-horizon inference~\cite{di2025streaming,kim2025infinipot,ning2025livevlm,
zhang2025flash} and proactive interaction~\cite{chen2024videollm,qian2025dispider,
wang2025streambridge, yang2025livestar}{}{.}
However, most existing frameworks implicitly assume that (i) the user query is
well-specified and (ii) the visual evidence needed to answer is available in the
current context.

In practice, both assumptions can fail. Under a bounded active-context budget,
visual content relevant to a later query may be compressed or removed from the
active context. Users may also pose imperfect queries, including referential
ambiguity (\eg, ``where did I leave it?'') and false premises (\eg, asking about
an object that never appeared), which can prevent reliable grounding even when
related frames remain visible. Although these failures have different causes,
current models tend to answer by default, producing fluent but ungrounded
responses, a common form of hallucination~\cite{bai2024hallucination,liu2024survey,
wang2024haloquest}. A useful
streaming assistant should instead decide whether to answer immediately or ask
for clarification.
Clarification can identify an ambiguous referent or
correct a false premise.

Making this work in streaming environments raises two challenges. First,
{}{deciding whether to answer directly or request clarification must add little latency:}
established uncertainty estimates often require multiple samples or repeated forward passes,
which {}{can add substantial query-time latency}, while trained ambiguity detectors may
not transfer across backbones or domains~\cite{farquhar2024detecting,
kuhn2023semantic,kadavath2022language}. Second, a clarified
query may refer to video content that is no longer represented in the bounded
active context. A practical system therefore needs a low-overhead
uncertainty check and access to earlier video content for
{}{answering after clarification}.

To address these requirements, we present \textit{StreamClarify}, a
training-free framework that augments a frozen streaming VLM with two
{}{components. First, the model makes a binary pre-answer}
\texttt{[Yes]/[No]} judgment about whether
its answer would rely on observed video evidence rather than general knowledge
alone, and StreamClarify uses the entropy
of this judgment as a lightweight
uncertainty signal. If the entropy exceeds a fixed threshold, answer generation
is suspended and clarification is triggered. Second, hierarchical memory
supports {}{answering} after clarification. The system keeps a small active
window of recent frames at full detail, compresses older content to stay within
a fixed context budget, and maintains a searchable CPU-side archive of
historical frames. Once clarified user guidance is available, StreamClarify uses
it to search the archive, prefills the retrieved frames
alongside the active context, and {}{generates the answer}.

\noindent\textbf{Our contributions are:}
\begin{itemize}
    \item We propose \textit{StreamClarify}, a training-free proactive
    clarification framework for frozen streaming VLMs. It combines a pre-answer
    uncertainty check with hierarchical memory to suspend answer generation,
    request clarification, and retrieve historical frames before {}{answering}.

    \item We introduce \textit{StreamClarify-QA}, a diagnostic streaming benchmark of 1,837
    QA pairs spanning standard, referential-ambiguity, and unanswerable
    scenarios, designed to expose answer-by-default behavior under
    imperfect queries in causal streaming conditions.

    \item Experiments across three streaming VLM
    backbones show Macro accuracy gains of up to 8.2 points over
    backbone-matched baselines while operating in real time. Additional
    experiments analyze the pre-answer uncertainty check, clarification
    effectiveness, and active-context management.
\end{itemize}

\section{Related Work}

\subsection{Streaming Video Understanding}
Unlike offline video understanding that processes pre-segmented clips in
batches~\cite{fu2025video, wu2024longvideobench, wang2025lvbench}, streaming
architectures process continuous frames in a single online pass. These approaches
broadly follow two directions. The first focuses on memory-efficient inference
{}{under a fixed active-context budget:} recent works manage the active context through
attention sinks~\cite{xu2025streamingvlm}, compress representations via token
merging or summarization~\cite{zhang2025flash, kim2025infinipot, tao2025dycoke,
xiong2025streaming,chen2025streamingtom}, or offload historical representations to CPU memory for
on-demand retrieval~\cite{ning2025livevlm, di2025streaming, kim2025v,
chen2025streamkv, patel2025cacheflow, zhang2026weavetime}, drawing on KV cache techniques originally developed for
text-only LLMs~\cite{xiao2023efficient, yang2024pyramidinfer, caipyramidkv}.
The second direction investigates proactive interaction: models like
VideoLLM-Online~\cite{chen2024videollm}, LiveStar~\cite{yang2025livestar},
and StreamMind~\cite{ding2025streammind} autonomously determine response
timing, while other frameworks adapt offline VLMs via plug-and-play
modules~\cite{wang2025streambridge}, asynchronous
pipelines~\cite{qian2025dispider}, or anticipatory
planning~\cite{yang2025streamagent} to enable continuous dialogue, accompanied
by emerging online benchmarks~\cite{lin2024streamingbench, niu2025ovo}.
Despite these architectural advances, current streaming systems generally
assume well-formed queries with fully observable visual evidence. {}{Our work focuses on how streaming models respond to imperfect queries.}

\subsection{Proactive Clarification and Uncertainty Estimation}
Proactive clarification has been explored in dialogue and static visual QA
via uncertainty-guided timing~\cite{testoni2024asking}, future-turn
anticipation~\cite{zhangmodeling}, and supervised policies for resolving
visual ambiguity~\cite{jian2025teaching}, but all of these require
task-specific training that limits their applicability to new domains or
backbone models. A related line of work examines whether VLMs can correctly
avoid answering when questions are unanswerable or based on false
premises~\cite{whitehead2022reliable, he2024tubench}, and hallucination
surveys have documented the prevalence of overconfident responses in
VLMs~\cite{bai2024hallucination, liu2024survey, wang2024haloquest}.
{}{These studies show that deciding when to suspend answering remains a
challenge for VLMs.}

{}
{For uncertainty estimation, training-free methods including semantic
entropy~\cite{kuhn2023semantic,farquhar2024detecting} and verbalized
self-assessment~\cite{kadavath2022language} avoid task-specific supervision but
may require multiple answer samples to estimate uncertainty.
Repeated sampling adds latency, which can limit its use in streaming QA.
Building on these ideas, we study uncertainty-guided clarification for imperfect
queries in streaming video QA without task-specific training.}

\begin{figure*}[t]
\centering
\includegraphics[width=0.95\textwidth]{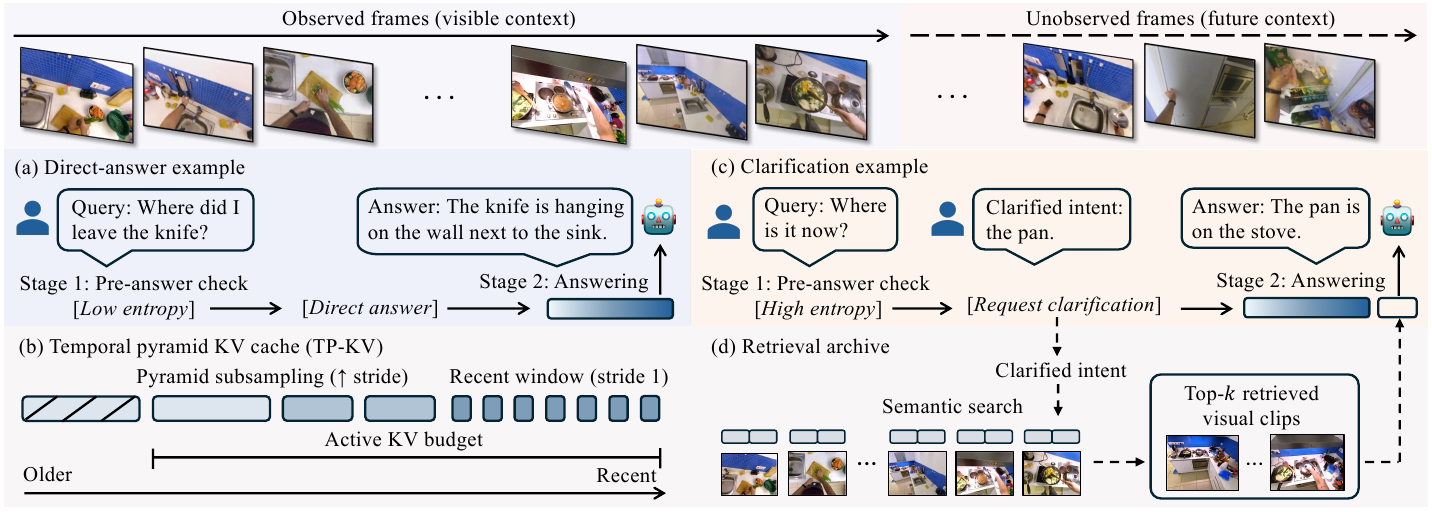}
\caption{Overview of StreamClarify. The framework augments a frozen streaming VLM with a pre-answer
uncertainty check in a two-stage generation loop and hierarchical memory.
(a)~When pre-answer
{}{entropy does not exceed the trigger threshold,} the system answers directly using the
active temporal-pyramid KV (TP-KV) cache. (b)~TP-KV keeps recent frames at full
resolution and compresses older observed frames within a fixed active budget.
(c)~When entropy exceeds the threshold, the system requests clarification and
converts the resulting user guidance into a retrieval query. (d)~Semantic
search retrieves relevant past clips from the archive and combines them with
the active TP-KV context for {}{answering after clarification}.}
\label{fig:main_figure}
\end{figure*}

\section{Methodology}

\subsection{\texorpdfstring{Problem Setup}{Problem Setup}}

{}{We model streaming video QA as a sequence of interactions}
$\langle V_i\rangle\langle Q_i\rangle\langle A_i\rangle$ ($i=1,2,\ldots$)
over a continuous stream $\mathcal{V}=\{F_1,F_2,\ldots\}$,
where $\langle V_i\rangle = \mathcal{V}_{\leq t_i}$ denotes the video
frames observed up to query time $t_i$, $\langle Q_i\rangle$ is the user
query, and $\langle A_i\rangle$ is the model response at round $i$.
At query time $t_i$, the model has access only to a
memory-constrained representation of $\langle V_i \rangle$
under budget $\mathcal{M}$,
so historical frames may be compressed or evicted before a query arrives.
In practice, $\mathcal{M}$ is implemented as a token budget $L_{\text{max}}$
for the active KV cache of the streaming backbone.

Beyond memory constraints, users may also issue \textit{imperfect queries},
such as queries with referential ambiguity or false premises.
Standard streaming VLMs lack an explicit mechanism to recognize when such
queries cannot be grounded.
In this work, the core
framework uses clarified user guidance and does not depend on a particular
question generator.
When clarification is requested, {}{the model delays
its answer $\langle A_i\rangle$ and obtains clarified user guidance}
$\langle R_i\rangle$. The model uses $\langle R_i\rangle$ to guide archive
retrieval before {}{generating its answer}. Thus, the interaction is extended from
$\cdots\langle V_i\rangle\langle Q_i\rangle\langle A_i\rangle\cdots$ to
$\cdots\langle V_i\rangle\langle Q_i\rangle
\langle R_i\rangle\langle A_i\rangle\cdots$.

\subsection{StreamClarify}

As shown in Fig.~\ref{fig:main_figure}, StreamClarify augments a frozen
streaming VLM with two training-free components: a pre-answer uncertainty check
for deciding whether to answer directly or request clarification, and a
hierarchical memory that maintains a bounded active context while providing
access to historical frames.

\subsubsection{\texorpdfstring{Pre-Answer Uncertainty Check}{Pre-Answer Uncertainty Check}}
\label{sec:evidence-check}

StreamClarify implements the pre-answer uncertainty check as a
two-stage generation pipeline that shares the same KV cache. Stage~1 asks
whether the answer would rely on observed video content rather than general
knowledge and emits a binary \texttt{[Yes]/[No]} judgment. Stage~2 generates
the answer conditioned on that judgment.

We measure uncertainty from the resulting two-label distribution. Let
$p$ denote the probability assigned to \texttt{[Yes]} at the token position
where the backbone distinguishes the labels. The binary entropy is
$H=-p\log_2(p)-(1-p)\log_2(1-p)$. If $H$ exceeds a fixed per-backbone threshold
set from the no-clarification distribution, answer generation is suspended and
clarification is triggered. This requires only one short generation before the
answer and no additional training. Supplementary Sec.~\ref*{sec:appendix-prompts} lists
the prompt templates.

\subsubsection{Hierarchical Memory}
\label{sec:hierarchical-memory}

Hierarchical memory combines a bounded temporal-pyramid KV cache
(TP-KV) with a CPU-side retrieval archive. Direct answers use only TP-KV. After
clarification, the user guidance queries the archive for frames no longer
retained in the active context.

\paragraph{TP-KV for bounded active context.}

TP-KV keeps recent frames at full granularity and compresses older history with a
temporal pyramid. Each incoming frame is first converted into backbone vision
tokens and prefilled through the model to update the active KV cache. When the
cache exceeds the token budget $L_{\text{max}}$, TP-KV protects a recent window,
compresses older tokens by stride sampling at progressively coarser pyramid
{}{levels, and evicts the oldest tokens if the compressed history still
exceeds the budget set by the compression ratio $\rho$.}
The retained tokens are then ordered chronologically and prefilled to rebuild a
bounded cache. This preserves recent detail while compactly representing earlier frames.

\vspace{-1.2em}
\paragraph{Retrieval archive for historical frames.}

For each incoming frame, the archive stores backbone vision tokens for
later prefilling and {}{a SigLIP2 image-text key~\cite{tschannen2025siglip}} for semantic search. After
{}{clarification, a search-query prompt converts the user guidance into visual
search phrases.} The top-$K$ historical clips are then prefilled through the
backbone and combined with the current TP-KV context before {}{answer generation}.
Supplementary Algorithm~\ref*{alg:memory} in Sec.~\ref*{sec:appendix-memory-algorithm}
details the online TP-KV update and archive construction procedure.
The
archive grows with streamed time and is measured separately from the bounded
active context. For longer streams, a retention window can
bound archive storage without changing the TP-KV budget.
Supplementary Sec.~\ref*{sec:appendix-memory-scalability}
{}{measures archive growth on an approximately 30-minute stream.}

\begin{figure}[t]
\centering
\includegraphics[width=\linewidth]{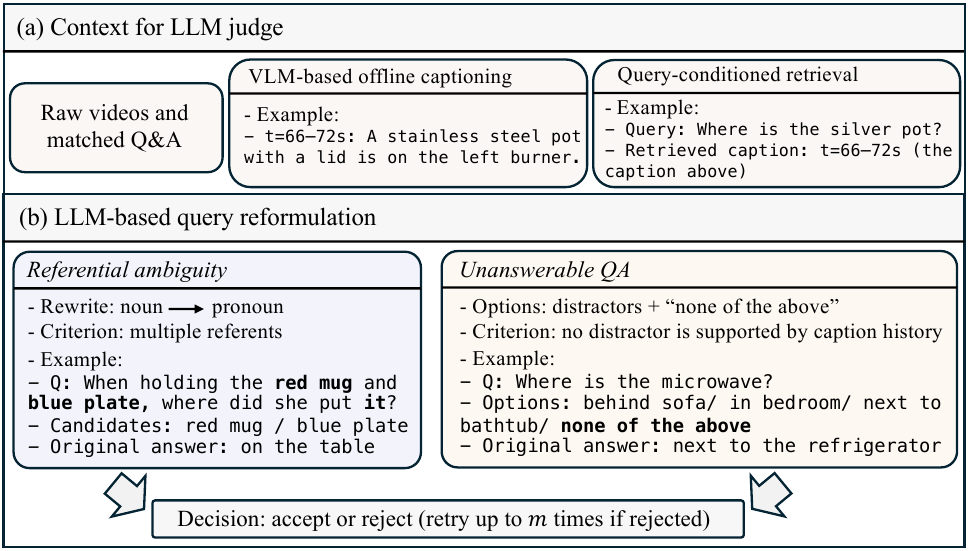}
\caption{StreamClarify-QA construction pipeline. (a) Dense captions are
aligned to each query timestamp. (b) Eligible QA pairs are transformed into
Referential Ambiguity or Unanswerable variants, followed by type-specific
validation. OVO-Bench HLD questions are retained unchanged.}
\label{fig:dataset}
\end{figure}

\section{\texorpdfstring{StreamClarify-QA Construction}{StreamClarify-QA Construction}}

To systematically evaluate model behavior under imperfect queries, we construct \textit{StreamClarify-QA}, a diagnostic streaming benchmark covering three scenario types. \textit{Standard QA} contains well-specified answerable queries. \textit{Referential Ambiguity QA} covers queries with underspecified references. \textit{Unanswerable QA} requires the model to select an abstention response because the observed video prefix supports none of the provided substantive answer options. The construction pipeline is illustrated in Fig.~\ref{fig:dataset}.

\subsection{Data Sources}
\label{sec:data-sources}
This diagnostic benchmark is built on two complementary datasets, summarized in Table~\ref{tab:dataset-stats}(a). \textit{StreamBench}{~\cite{xiong2025streaming}} contributes 1,556 deduplicated QA pairs from 260 videos (avg.\ 253~s) in open-ended format. \textit{OVO-Bench} provides questions in multiple-choice format. We select all videos containing at least one Hallucination Detection (HLD) question, together with all other questions from those videos. OVO-Bench defines HLD as questions irrelevant to the available video input~\cite{niu2025ovo}. {}{To avoid closely spaced queries,} questions with a temporal gap of less than 5 seconds from the preceding question are removed, retaining 281 QA pairs from 95 videos (avg.\ 383~s). All questions are evaluated causally at their annotated timestamps, so the model can use only the video prefix available when each question arrives.

\subsection{Query Transformation}

{}{MiniCPM-V 4.5~\cite{yu2025minicpm} generates
timestamp-aligned dense captions conditioned on source questions but not answers
(Supplementary Sec.~\ref*{sec:appendix-uq-audit}).}
{}{General-knowledge questions}
are excluded{ from transformation}, leaving 1,302 eligible pairs from
StreamBench and 130 non-HLD pairs from OVO-Bench. Unanswerable construction
additionally excludes 307 yes/no StreamBench questions, leaving 995 source
pairs. All transformations are
performed using Qwen3-14B~\cite{yang2025qwen3}.

\noindent\textbf{Referential Ambiguity.} An LLM rewrites each eligible query by replacing a specific noun phrase
{}{with an ambiguous pronoun, provided that the captions contain multiple
plausible referents at query time.} Invalid rewrites are
retried up to 3 times. This yields 351 and 53 candidate pairs from
StreamBench and OVO-Bench respectively, before human quality filtering.
Pairs that exhaust their retry budget are preserved as Standard QA.

\noindent\textbf{Unanswerable QA.} Inspired by the native HLD category in
OVO-Bench~\cite{niu2025ovo}, we construct a complementary Unanswerable split
for StreamBench and combine the two sources in StreamClarify-QA. Each
remaining
open-ended StreamBench question is converted into multiple-choice format by
generating plausible distractors. Each candidate set is checked against the
caption history up to the query timestamp. If any distractor is supported by
the captions, the set is rejected and regenerated for up to three attempts.
After validation, the source answer is removed and ``None of the above'' is
added. The 151 OVO-Bench HLD questions are retained unchanged, while the 389
constructed StreamBench instances undergo caption-based filtering and
raw-video label validation in Sec.~\ref{sec:quality-control}. Together, they
form 540 Unanswerable QA pairs.

Transformations are applied independently to individual QA pairs while
preserving their original timestamps in the continuous video stream. OVO-Bench
does not annotate linked dialogue turns, while StreamBench transformations do
not deliberately select or link earlier QA turns that identify the intended
referent. We therefore evaluate clarification of the current query without
treating prior dialogue as a defined benchmark input.

\subsection{Quality Control and Dataset Statistics}
\label{sec:quality-control}

Timestamp-aligned captions guide both query transformations, followed by
query-specific quality checks. Because ambiguity depends on the final wording,
human review of all 404 Referential Ambiguity candidates removes insufficiently
ambiguous rewrites and cases whose referent remains recoverable from the
original annotations, leaving 302 pairs.
For StreamBench Unanswerable QA, caption-based filtering and balanced split
assembly retain 389 instances. To account for visual information that captions
may omit, we screen all 389 against the causal video prefix available at the
query timestamp. Three rounds of Gemini~\cite{google2026gemini36flash}
screening leave seven cases for human review, which confirms the original label
in every case. The 151 OVO-Bench HLD questions are retained unchanged and are
not part of this audit. Supplementary Sec.~\ref*{sec:appendix-uq-audit} reports the
filtering counts and raw-video protocol.
The final diagnostic benchmark comprises 1,837 QA pairs across 355 videos: 995
Standard, 302 Referential Ambiguity, and 540 Unanswerable QA instances,
{}{as detailed in Tab.~\ref{tab:dataset-stats}(b). Imperfect-query scenarios
account for nearly half of all instances, allowing evaluation beyond standard QA.}

\begin{table}[t]
\centering
\caption{Statistics of StreamClarify-QA. (a) Video and QA overview by
source. (b)~Scenario distribution across both datasets. Abbreviations: Vid =
videos, Dur. = duration, Ref. Ambig. = referential ambiguity, and Unans. =
unanswerable.}
\label{tab:dataset-stats}
\setlength{\tabcolsep}{0.9pt}
\renewcommand{\arraystretch}{0.94}
\scriptsize
\begin{minipage}[t]{0.53\linewidth}
\centering
\textbf{(a)} Overview\\[1pt]
\begin{tabular}[t]{@{}lrrrr@{}}
\toprule
Dataset & \#Vid & \#QA & Q/Vid & Dur.(s) \\
\midrule
StreamBench & 260 & 1{,}556 & 6.0 & ${\sim}$253 \\
OVO-Bench  &  95 &    281 & ${\sim}$3.0 & ${\sim}$383 \\
\bottomrule
\end{tabular}
\end{minipage}\hfill
\begin{minipage}[t]{0.45\linewidth}
\centering
\textbf{(b)} Scenario distribution\\[1pt]
\begin{tabular}[t]{@{}lrrr@{}}
\toprule
Scenario & Total & Stream. & OVO \\
\midrule
A. Standard    & 995 & 913 &  82 \\
B. Ref. Ambig. & 302 & 254 &  48 \\
C. Unans. QA   & 540 & 389 & 151 \\
\bottomrule
\end{tabular}
\end{minipage}
\end{table}

\section{Experiments}

\subsection{Settings}

\noindent\textbf{Baselines.}
Baselines include closed-source, training-based, and backbone-matched
training-free systems. The closed-source reference is
GPT-5-mini~\cite{singh2025openai} evaluated {}{on recent frames at each query}. Training-based baselines include
VideoLLM-Online~\cite{chen2024videollm}, Flash-VStream~\cite{zhang2025flash},
ReKV~\cite{di2025streaming}, and TimeChat{-Online}~\cite{yao2025timechat}.
{}{For the backbone-matched comparison,
we apply the memory mechanisms of} StreamChat~\cite{xiong2025streaming},
StreamBridge~\cite{wang2025streambridge}, and
StreamingVLM~\cite{xu2025streamingvlm}{}{ to} the same three
{frozen }backbones as StreamClarify{, without additional training}.

\noindent\textbf{Evaluation Metrics.} We report Micro Accuracy over all QA, Macro Accuracy averaged across
scenario types, time-to-first-token (TTFT), and online processing throughput
{}{(ProcFPS). For trigger analyses, we report the trigger rate and measure
Precision and Recall with incorrect initial answers as positives. Error AUROC
measures error ranking across thresholds.} Clarification effects are reported through
wrong-to-correct and correct-to-wrong transitions. The main, matched-6K, and
entropy-gated Generated Clarification comparisons include 95\% confidence intervals
from paired resampling of complete videos. Correctness is judged by
Qwen3-14B~\cite{yang2025qwen3} using dense offline captions as scene-level
reference context.

\noindent\textbf{Implementation Details.} StreamClarify and the backbone-matched training-free baselines use
three frozen VLMs: LLaVA-OV-1.5~{}{\cite{an2025llava15}},
InternVL-3.5~\cite{wang2025internvl3}, and Qwen3-VL~\cite{bai2025qwen3}, without
task-specific training. The main comparison follows each method's standard
configuration. StreamClarify uses a 10K-token TP-KV budget, compression ratio
0.25, and a 20~s recency window. We also compare all Qwen3-VL methods under a
common 6K active-context budget, excluding CPU-side storage. Clarification uses
a per-backbone threshold {}{set from the pooled evaluation's no-clarification entropy distribution, without a separate calibration split,} to obtain a
near-30\% rate over the pooled $1{,}837$-QA evaluation and retrieves the
top-$K{=}16$ clips.
The evaluation population, backbone, and clarification setting for each
analysis are summarized in
Supplementary Table~\ref*{tab:evaluation-scope}.

\subsection{Main Results}

\begin{table*}[t]
\centering
\caption{Per-category streaming video QA results on StreamBench and OVO-Bench
in the no-clarification setting using each method's standard
configuration. Shaded columns denote the overall accuracy and
system-efficiency columns. Greyed text reports the {}per-category breakdown
(Std.\ $=$ Standard, Amb.\ $=$ Ambiguous, UQ $=$ Unanswerable QA). Arrows indicate
the preferred direction. Accuracy values are percentages, and TTFT is
reported in seconds.}
\label{tab:main-comparison}
\setlength{\tabcolsep}{3.4pt}
\renewcommand{\arraystretch}{0.94}
\footnotesize
\begin{tabular}{@{}l@{\hspace{3pt}}l rrr
    >{\columncolor{blue!7}}r >{\columncolor{blue!7}}r
    >{\columncolor{gray!12}}r >{\columncolor{gray!12}}r
  !{\hspace{7pt}}
  rrr
    >{\columncolor{blue!7}}r >{\columncolor{blue!7}}r
    >{\columncolor{gray!12}}r >{\columncolor{gray!12}}r
  @{}}
\toprule
& & \multicolumn{7}{c}{StreamBench}
& \multicolumn{7}{c}{OVO-Bench} \\
\cmidrule(lr){3-9}\cmidrule(l){10-16}
& & \multicolumn{3}{c}{\textcolor{black!40}{Per-Cat.\ Acc.}}
& \multicolumn{2}{>{\columncolor{blue!7}}c}{\textbf{Overall Acc.}}
& \multicolumn{2}{>{\columncolor{gray!12}}c}{\textbf{System}}
& \multicolumn{3}{c}{\textcolor{black!40}{Per-Cat.\ Acc.}}
& \multicolumn{2}{>{\columncolor{blue!7}}c}{\textbf{Overall Acc.}}
& \multicolumn{2}{>{\columncolor{gray!12}}c}{\textbf{System}} \\
\cmidrule(lr){3-5}\cmidrule(lr){6-7}\cmidrule(lr){8-9}
\cmidrule(lr){10-12}\cmidrule(lr){13-14}\cmidrule(l){15-16}
& & \textcolor{black!40}{Std.}
& \textcolor{black!40}{Amb.}
& \textcolor{black!40}{UQ}
& \cellcolor{blue!7}\textbf{Micro}
& \cellcolor{blue!7}\textbf{Macro}
& \cellcolor{gray!12}TTFT$\downarrow$
& \cellcolor{gray!12}ProcFPS$\uparrow$
& \textcolor{black!40}{Std.}
& \textcolor{black!40}{Amb.}
& \textcolor{black!40}{UQ}
& \cellcolor{blue!7}\textbf{Micro}
& \cellcolor{blue!7}\textbf{Macro}
& \cellcolor{gray!12}TTFT$\downarrow$
& \cellcolor{gray!12}ProcFPS$\uparrow$ \\
\midrule
\scriptsize\emph{Closed} & GPT-5-mini
  & \textcolor{black!45}{73.4} & \textcolor{black!45}{16.9} & \textcolor{black!45}{ 8.0}
  & 47.8 & 32.8 & {--} & {--}
  & \textcolor{black!45}{58.5}
  & \textcolor{black!45}{54.2}
  & \textcolor{black!45}{25.2}
  & 39.9
  & 46.0
  & 3.829
  & {--} \\
\midrule
\multirow{4}{*}{\scriptsize\emph{Open}} & VideoLLM-Online
  & \textcolor{black!45}{49.5} & \textcolor{black!45}{11.8} & \textcolor{black!45}{ 0.5}
  & 31.1 & 20.6 & 0.064 &  9.43
  & \textcolor{black!45}{13.4} & \textcolor{black!45}{22.9} & \textcolor{black!45}{ 0.7}
  &  8.2 & 12.3 & 0.066 & 11.44 \\
& Flash-VStream
  & \textcolor{black!45}{62.1} & \textcolor{black!45}{11.4} & \textcolor{black!45}{17.2}
  & 42.6 & 30.2 & 0.149 &  1.97
  & \textcolor{black!45}{45.1}
  & \textcolor{black!45}{47.9}
  & \textcolor{black!45}{16.6}
  & 30.2
  & 36.5
  & 0.125 &  1.91 \\
& ReKV
  & \textcolor{black!45}{67.1} & \textcolor{black!45}{16.5} & \textcolor{black!45}{ 6.2}
  & 43.6 & 29.9 & 0.181 &  4.53
  & \textcolor{black!45}{58.5}
  & \textcolor{black!45}{47.9}
  & \textcolor{black!45}{30.5}
  & 41.6
  & 45.6
  & 0.053 &  6.76 \\
& TimeChat{-Online}
  & \textcolor{black!45}{64.4} & \textcolor{black!45}{14.2} & \textcolor{black!45}{29.3}
  & 47.4 & 36.0 & 2.045 &  5.80
  & \textcolor{black!45}{45.1}
  & \textcolor{black!45}{37.5}
  & \textcolor{black!45}{37.1}
  & 39.5
  & 39.9
  & 1.481
  & 7.60 \\
\midrule
\multirow{4}{*}{\scriptsize\emph{LLaVA-OV}} & StreamChat
  & \textcolor{black!45}{64.0} & \textcolor{black!45}{15.4} & \textcolor{black!45}{46.5}
  & 51.7 & 41.9 & 0.700 &  2.28
  & \textcolor{black!45}{53.7} & \textcolor{black!45}{35.4} & \textcolor{black!45}{66.9}
  & 57.7 & 52.0 & 0.642 &  2.45 \\
& StreamBridge
  & \textcolor{black!45}{73.1} & \textcolor{black!45}{21.7} & \textcolor{black!45}{19.0}
  & 51.2 & 37.9 & 1.673 &  4.80
  & \textcolor{black!45}{68.3} & \textcolor{black!45}{43.8} & \textcolor{black!45}{44.4}
  & 51.2 & 52.1 & 2.137 &  7.46 \\
& StreamingVLM
  & \textcolor{black!45}{70.0} & \textcolor{black!45}{22.0} & \textcolor{black!45}{26.2}
  & 51.2 & 39.4 & 0.092 &  1.79
  & \textcolor{black!45}{61.0} & \textcolor{black!45}{47.9} & \textcolor{black!45}{43.0}
  & 49.1 & 50.6 & 0.113 &  1.80 \\
& Ours
  & \textcolor{black!45}{64.5} & \textcolor{black!45}{16.5} & \textcolor{black!45}{45.2}
  & \bf51.9 & \bf42.1 & 0.313 & 4.98
  & \textcolor{black!45}{64.6} & \textcolor{black!45}{47.9} & \textcolor{black!45}{68.2}
  & \bf63.7 & \bf60.3 & 0.334 & 5.57 \\
\midrule
\multirow{4}{*}{\scriptsize\emph{InternVL}} & StreamChat
  & \textcolor{black!45}{63.5} & \textcolor{black!45}{14.6} & \textcolor{black!45}{40.1}
  & 49.7 & 39.4 & 0.594 &  3.25
  & \textcolor{black!45}{52.4} & \textcolor{black!45}{39.6} & \textcolor{black!45}{36.4}
  & 41.6 & 42.8 & 0.514 &  4.06 \\
& StreamBridge
  & \textcolor{black!45}{74.3} & \textcolor{black!45}{19.3} & \textcolor{black!45}{39.1}
  & \bf56.5 & 44.2 & 2.894 &  9.43
  & \textcolor{black!45}{57.3} & \textcolor{black!45}{41.7} & \textcolor{black!45}{49.7}
  & 50.5 & 49.6 & 3.056 & 26.95 \\
& StreamingVLM
  & \textcolor{black!45}{71.1} & \textcolor{black!45}{16.1} & \textcolor{black!45}{26.2}
  & 50.9 & 37.8 & 0.072 &  3.08
  & \textcolor{black!45}{52.4} & \textcolor{black!45}{43.8} & \textcolor{black!45}{19.9}
  & 33.5 & 38.7 & 0.078 &  3.20 \\
& Ours
  & \textcolor{black!45}{67.6} & \textcolor{black!45}{14.6} & \textcolor{black!45}{53.7}
  & 55.5 & \bf45.3 & 0.240 & 5.53
  & \textcolor{black!45}{59.8} & \textcolor{black!45}{50.0} & \textcolor{black!45}{56.3}
  & \bf56.2 & \bf55.3 & 0.258 & 5.89 \\
\midrule
\multirow{4}{*}{\scriptsize\emph{Qwen3-VL}} & StreamChat
  & \textcolor{black!45}{60.0} & \textcolor{black!45}{16.1} & \textcolor{black!45}{62.7}
  & 53.5 & 46.3 & 0.305 &  3.50
  & \textcolor{black!45}{45.1} & \textcolor{black!45}{33.3} & \textcolor{black!45}{58.3}
  & 50.2 & 45.6 & 0.191 &  3.79 \\
& StreamBridge
  & \textcolor{black!45}{75.7} & \textcolor{black!45}{23.6} & \textcolor{black!45}{35.7}
  & 57.2 & 45.0 & 2.657 & 16.68
  & \textcolor{black!45}{56.1} & \textcolor{black!45}{43.8} & \textcolor{black!45}{37.7}
  & 44.1 & 45.9 & 2.857 & 36.93 \\
& StreamingVLM
  & \textcolor{black!45}{68.3} & \textcolor{black!45}{21.7} & \textcolor{black!45}{54.0}
  & 57.1 & 48.0 & 0.062 &  8.70
  & \textcolor{black!45}{59.8} & \textcolor{black!45}{33.3} & \textcolor{black!45}{43.0}
  & 46.3 & 45.4 & 0.069 &  9.33 \\
& Ours
  & \textcolor{black!45}{67.4} & \textcolor{black!45}{19.3} & \textcolor{black!45}{67.4}
  & \bf59.5 & \bf51.3 & 0.239 & 10.16
  & \textcolor{black!45}{58.5} & \textcolor{black!45}{39.6} & \textcolor{black!45}{62.9}
  & \bf57.7 & \bf53.7 & 0.260 & 12.38 \\
\bottomrule
\end{tabular}
\end{table*}

\paragraph{Overall Performance.}

Table~\ref{tab:main-comparison} shows that StreamClarify obtains the
highest Macro accuracy in all six dataset--backbone settings and the highest
{}{Micro accuracy in five. Compared with the strongest matched baseline for
each metric, Qwen3-VL gains 2.3 Micro and 3.3 Macro points on StreamBench.} Under InternVL-3.5, Micro accuracy is 1.0 point below
StreamBridge, while Macro accuracy is 1.1 points higher. On OVO-Bench, the
largest gains are 7.5 points in Micro under Qwen3-VL and 8.2 points in Macro
under LLaVA-OV-1.5. Paired video-level intervals exclude zero for the Qwen3-VL Macro
gain on StreamBench ($+3.3$ points, 95\% CI $[+1.0,+5.7]$) and for the
LLaVA-OV-1.5 and Qwen3-VL Macro gains on OVO-Bench ($+8.2$ points,
$[+2.2,+13.8]$, and $+7.8$ points, $[+2.4,+12.9]$). Supplementary Sec.~\ref*{sec:appendix-main-ci}
reports the remaining intervals. Against {}TimeChat{-Online}, Qwen3-VL StreamClarify
gains 12.1 Micro and 15.3 Macro points on StreamBench.

The per-category results in Table~\ref{tab:main-comparison} show that
the clearest gains occur on Unanswerable QA. On OVO-Bench, StreamClarify reaches
68.2\%, 56.3\%, and 62.9\% across the three backbones, improving over their
strongest matched baselines by 1.3, 6.6, and 4.6 points. Standard and Ambiguous
QA vary more across methods, motivating the clarification analysis in
Sec.~\ref{sec:clarification}. The {} GPT-5-mini reference also drops from
73.4\% on Standard QA to 16.9\% on Ambiguous and 8.0\% on Unanswerable QA on
StreamBench, indicating that general model capability alone does not resolve
imperfect queries.

\paragraph{System Efficiency.}

Table~\ref{tab:main-comparison} reports efficiency in the same
no-clarification setting. With Qwen3-VL on StreamBench, StreamClarify processes
10.2~FPS with 0.239~s TTFT, compared with 3.5~FPS and 0.305~s for StreamChat.
StreamBridge processes 16.7~FPS but has 2.657~s TTFT, while StreamingVLM has
0.062~s TTFT and retains only its attention sinks and recent window in the
active visual context. Clarification-enabled latency is evaluated separately in
Sec.~\ref{sec:clarification}.

\subsection{Effectiveness of Proactive Clarification}
\label{sec:clarification}
We evaluate proactive clarification at two levels. The oracle-guided
analysis evaluates the entropy trigger together with archive retrieval and
re-answering across three
question types and three backbones. We then report a generated-question proof
of concept on Ambiguous queries with simulated referent-only replies.

\paragraph{Oracle-Guided Analysis.}

Table~\ref{tab:clarification-effectiveness} evaluates three backbones
with oracle guidance, providing a controlled upper bound on user-guidance
{}{quality. For
Referential Ambiguity QA, the guidance supplies the correct referent. For Standard
and transformed Unanswerable QA, we derive user guidance from the original
pre-transformation instance.} Naturally unanswerable queries use the question
itself as the retrieval query. These inputs are used only after a trigger and do
not change the initial active KV cache or insert the reference answer into the
answering context. This analysis tests whether the trigger ranks
initial-answer errors and whether the system can use correct guidance
when it is available.

\begin{table}[!t]
\centering
\caption{Oracle-guided analysis on StreamBench and OVO-Bench pooled
{} at comparable trigger
rates near 30\%. Guidance is provided only after a trigger. Grey text
shows per-category accuracy, while shaded columns report overall accuracy and
{}{efficiency. Trig. is the trigger rate; Prec. and Rec. treat initially
incorrect answers as positives.}
{}{TTFT is averaged
over all 1,837 queries. ProcFPS is the unweighted mean of per-video processing
throughput across 355 videos.}}
\label{tab:clarification-effectiveness}
\setlength{\tabcolsep}{1.4pt}
\renewcommand{\arraystretch}{0.96}
\scriptsize
\resizebox{\linewidth}{!}{%
\begin{tabular}{@{}l rrr
    >{\columncolor{blue!7}}r >{\columncolor{blue!7}}r
    >{\columncolor{gray!12}}r >{\columncolor{gray!12}}r
  @{\hspace{3pt}}
  rrr@{}}
\toprule
& \multicolumn{7}{c}{Overall Performance}
& \multicolumn{3}{c}{Clarification} \\
\cmidrule(lr){2-8}\cmidrule(l){9-11}
Method
& \textcolor{black!40}{Std.}
& \textcolor{black!40}{Amb.}
& \textcolor{black!40}{UQ}
& \cellcolor{blue!7}\textbf{Micro}
& \cellcolor{blue!7}\textbf{Macro}
& \cellcolor{gray!12}TTFT$\downarrow$
& \cellcolor{gray!12}ProcFPS$\uparrow$
& {Trig.} & {Prec.} & {Rec.} \\
\midrule
\multicolumn{11}{@{}l}{\emph{LLaVA-OV-1.5}}\\
w/o Clar.
  & \textcolor{black!45}{64.5} & \textcolor{black!45}{21.5} & \textcolor{black!45}{51.7}
  & 53.7 & 45.9 & 0.316 & 5.14
  & 0.0 & {--} & {--} \\
w/ Random
  & \textcolor{black!45}{65.6} & \textcolor{black!45}{24.8} & \textcolor{black!45}{49.3}
  & 54.1 & 46.6 & 1.145 & 4.05
  & 27.8 & 44.5
  & 26.7 \\
w/ Entropy
  & \textcolor{black!45}{66.9} & \textcolor{black!45}{31.1} & \textcolor{black!45}{49.8}
  & \bf56.0 & \bf49.3 & 1.166 & 4.04
  & 28.4 & \bf63.0
  & \bf38.5 \\
\midrule
\multicolumn{11}{@{}l}{\emph{InternVL-3.5}}\\
w/o Clar.
  & \textcolor{black!45}{66.9} & \textcolor{black!45}{20.2} & \textcolor{black!45}{54.4}
  & 55.6 & 47.2 & 0.243 & 5.63
  & 0.0 & {--} & {--} \\
w/ Random
  & \textcolor{black!45}{68.2} & \textcolor{black!45}{24.8} & \textcolor{black!45}{55.0}
  & 57.2 & 49.4 & 0.979 & 4.33
  & 31.3 & 43.7
  & 30.8 \\
w/ Entropy
  & \textcolor{black!45}{68.4} & \textcolor{black!45}{26.5} & \textcolor{black!45}{57.2}
  & \bf58.2 & \bf50.7 & 0.976 & 4.31
  & 30.9 & \bf56.7
  & \bf39.5 \\
\midrule
\multicolumn{11}{@{}l}{\emph{Qwen3-VL}}\\
w/o Clar.
  & \textcolor{black!45}{66.6} & \textcolor{black!45}{22.5} & \textcolor{black!45}{66.1}
  & 59.2 & 51.8 & 0.242 & 10.75
  & 0.0 & {--} & {--} \\
w/ Random
  & \textcolor{black!45}{67.0} & \textcolor{black!45}{28.1} & \textcolor{black!45}{66.1}
  & 60.4 & 53.8 & 1.030 & 7.57
  & 33.8 & 41.6
  & 34.4 \\
w/ Entropy
  & \textcolor{black!45}{66.8} & \textcolor{black!45}{30.8} & \textcolor{black!45}{68.1}
  & \bf61.3 & \bf55.3 & 1.019 & 7.51
  & 33.0 & \bf59.8
  & \bf48.5 \\
\bottomrule
\end{tabular}
}
\end{table}

Table~\ref{tab:clarification-effectiveness}
shows that entropy-based triggering yields higher Micro and Macro Accuracy than
{}{both no clarification and random triggering at comparable rates for all three}
backbones under oracle guidance. With Qwen3-VL, Micro accuracy is
61.3\%, compared with 59.2\% without clarification and 60.4\% with random
triggering. At comparable
trigger rates near 33\%, entropy obtains
59.8\% Precision and 48.5\%
Recall for initial-answer errors, compared with
41.6\% and 34.4\% for random.
These accuracy gains come with additional query-time cost. {}{In this oracle-guided setting, Qwen3-VL's
mean TTFT over all queries} increases from 0.242 to 1.019~s at the 33.0\% trigger rate.
Section~5.4 isolates the overhead of the pre-answer check, and
Supplementary Sec.~\ref*{sec:appendix-clarification-latency} {}{reports total model-side runtime
for all-query generated clarification}.

\paragraph{Generated-Question Proof of Concept.}

We further evaluate generated clarification on the 302 Ambiguous QA
pairs from StreamBench and OVO-Bench using Qwen3-VL and simulated referent-only
replies. We reuse the numerical threshold from the oracle-guided analysis
without retuning. It triggers on about one-third of the
pooled $1{,}837$-QA evaluation and on
162/302 (53.6\%) Ambiguous queries. Direct Retrieval,
Generated Clarification, and Oracle Guidance start from the same reconstructed
causal state and use the same retrieval and re-answering procedure.
{}{\textit{w/o Clar.} retains the original online answer.}
\textit{Direct Retrieval} uses the ambiguous query
without user guidance, \textit{Generated Clarification} asks one short
question, and \textit{Oracle Guidance} directly receives the intended
referent.

In the all-query Generated Clarification condition, question generation is
attempted without entropy selection. The generator receives the ambiguous query
and causal visual state, but not the intended referent, original unambiguous
question, reference answer, or future frames. An evaluation-only GPT-5
gate~\cite{singh2025openai} then checks each generated question against the
intended referent. It accepts 212
questions and routes the remaining 90 to Direct Retrieval. The same fallback
rule is applied within the entropy-gated condition. On accepted items, Generated
Clarification and Oracle Guidance receive the same fixed referent-only reply,
\texttt{I mean \{referent\}.}, and follow the same downstream retrieval and
re-answering path. Under entropy gating, this generation-and-gate procedure is
applied to the 162 selected queries, while the original answer is retained for
the remaining 140. {}{Random gating uniformly selects 162/302 queries over $1{,}000$ trials,
reusing the same per-query outcomes (Supplementary Sec.~\ref*{sec:appendix-random-generated}).}
Exact prompts
and gate details are provided in
Supplementary Sec.~\ref*{sec:appendix-generated-question-prompt}.

\begin{table}[!b]

\centering
\caption{Generated-question analysis on 302 Ambiguous QA pairs with Qwen3-VL.
{}
{Random-gated entries, including paired transition counts, are means over $1{,}000$ selections of 162 queries.} W$\rightarrow$C
and C$\rightarrow$W denote paired correctness transitions.}
\label{tab:generated-clarification}
\setlength{\tabcolsep}{1.6pt}
\renewcommand{\arraystretch}{0.95}
\scriptsize
\resizebox{\linewidth}{!}{%
\begin{tabular}{@{}l l r r r r r@{}}
\toprule
& & \multicolumn{2}{c}{Overall Accuracy}
& \multicolumn{3}{c}{Paired Change} \\
\cmidrule(lr){3-4}\cmidrule(l){5-7}
Condition
& Selection
& Correct
& Micro (\%)
& W$\rightarrow$C
& C$\rightarrow$W
& $\Delta$Correct \\
\midrule
w/o Clar. & None & 68 & 22.5 & -- & -- & -- \\
{Generated Clar.} & {Random-gated} & {87.8} & {29.1} & {29.5} & {9.7} & {+19.8} \\
Generated Clar. & Entropy-gated & \bf93 & \bf30.8 & \bf34 & \bf9 & \textbf{+25} \\
\midrule
Direct Retrieval & All-query & 69 & 22.8 & 20 & 19 & +1 \\
Generated Clar. & All-query & 105 & 34.8 & 55 & 18 & +37 \\
Oracle Guidance & All-query & 116 & 38.4 & 62 & 14 & +48 \\
\bottomrule
\end{tabular}}
\end{table}

As shown in Table~\ref{tab:generated-clarification}, entropy-gated
Generated Clarification improves accuracy from 22.5\% to
30.8\% ($+8.3$ points, 95\% CI $[+4.2,+12.5]$), {}{with 34 corrections and 9
new errors.} {With the same number of selected queries, random gating averages 29.1\%
accuracy, 1.7 points below entropy gating.} Direct Retrieval yields only one additional correct answer, showing that
retrieval and re-answering alone do not explain the gain. Without entropy
selection, Generated Clarification reaches 34.8\%, compared with the 38.4\%
{}{accuracy under correct-referent oracle guidance}.
The gated gain remains positive under separate blinded
GPT-5 and Gemini 3.6 Flash answer judges (Supplementary Sec.~\ref*{sec:appendix-evaluator-sensitivity}).
Detailed clarification latency is reported in
Supplementary Sec.~\ref*{sec:appendix-clarification-latency}.

\subsection{Analysis and Ablations}
We analyze pre-answer scoring, TP-KV, and clarification parameters.
The pre-answer scoring and TP-KV ablations use Qwen3-VL on OVO-Bench
without clarification or retrieval. The retrieval ablation uses the same dataset
and backbone under oracle user guidance. The matched-6K comparison instead
evaluates Qwen3-VL on the full StreamBench and OVO-Bench with
clarification disabled.

\paragraph{Pre-Answer Scoring Strategy.}

Table~\ref{tab:ablation_uncertainty} and Figure~\ref{fig:pr_curves} compare
our two-stage design with four alternatives. Direct Answer prompts the model to
answer immediately, without a separate pre-answer judgment. Ours (Single)
produces the binary judgment and answer in one generation, while Probability
requests an explicit confidence score and V/K Markers substitutes arbitrary
tokens for the semantic decision labels. Clarification, retrieval, and
re-answering are disabled. Table~\ref{tab:ablation_uncertainty}
reports the accuracy and latency of the resulting initial answers, while
Figure~\ref{fig:pr_curves} evaluates how well each score identifies errors in
those answers. {The error labels are method-specific, and threshold
selection yields different query counts across methods.}

\begin{table}[t]
\centering
\caption{Comparison of pre-answer scoring strategies on OVO-Bench with
Qwen3-VL without clarification or retrieval.}
\label{tab:ablation_uncertainty}
\scriptsize
\setlength{\tabcolsep}{8pt}
\renewcommand{\arraystretch}{0.95}
\begin{tabular}{@{}lcc@{}}
    \toprule
    Method & Macro (\%) & TTFT (s) \\
    \midrule
    Direct Answer           & 49.3 & 0.085 \\
    Probability      & 50.3 & 0.365 \\
    V/K Markers      & 51.3 & 0.210 \\
    Ours (Single)    & 51.0 & 0.100 \\
    \rowcolor{gray!10}
    Ours (Two-stage) & \bf53.7 & 0.260 \\
    \bottomrule
\end{tabular}
\end{table}

\begin{figure}[t]
\centering
\includegraphics[width=\linewidth]{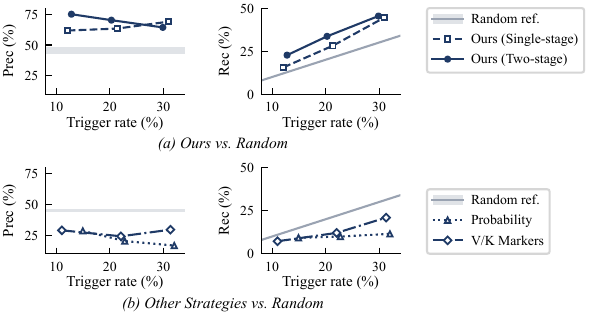}
\caption{{}{Trigger Precision and Recall on OVO-Bench with Qwen3-VL.
Trigger rates are computed over all 281 questions. Gray bands span the methods'
expected random Precision (a: 42.35--48.04\%, b: 44.13--46.26\%).
They are not confidence intervals.
Gray lines show expected Recall under uniform random selection, equal to the
trigger rate.}}
\label{fig:pr_curves}
\end{figure}

Our two-stage design reaches 53.7\% Macro Accuracy on its
initial answers, 2.4 points above the strongest alternative
(Table~\ref{tab:ablation_uncertainty}). {}{The single-stage and
two-stage variants exceed their random Precision and Recall expectations at
all reported points}
(Figure~\ref{fig:pr_curves}a). The additional pre-answer pass increases TTFT
from 0.085 to 0.260~s. Probability and V/K Markers fall below random triggering
in Figure~\ref{fig:pr_curves}b, supporting the use of semantically meaningful
decision labels instead of arbitrary markers or an explicitly generated
probability. {}
{}
{For the same \texttt{[Yes]/[No]} judgment, entropy ranks answer errors
better than $p(\mathrm{No})$ on StreamBench. Its ranking performance is also
stable under the tested prompt perturbations
(Supplementary Sec.~\ref*{sec:appendix-trigger-robustness}).}

\begin{figure}[h]
\centering
\includegraphics[width=\linewidth]{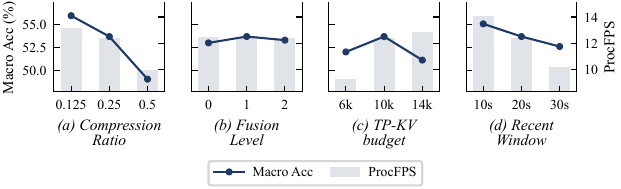}
\caption{TP-KV hyperparameter ablations on OVO-Bench with Qwen3-VL in the
no-clarification setting. Bars show ProcFPS and lines show
Macro Accuracy.}
\label{fig:ablation_tpkv}
\end{figure}

\paragraph{TP-KV Memory Hyperparameters.}
{}{We vary four hyperparameters that control how tokens are}
{}{compressed (compression ratio and fusion level) and how the budget is}
allocated (budget size and
recency window size). {Fusion Level sets the maximum level reached
when merging historical clip representations. The default is 1. Level 0
disables fusion but still allows eviction.}
The results in
Figure~\ref{fig:ablation_tpkv} show that a lower compression ratio
improves both Macro Accuracy and ProcFPS, while performance varies only
slightly across fusion levels. The budget ablation favors the middle 10K
setting, and shorter recency windows allocate more capacity to compressed
history while improving Macro Accuracy.

\paragraph{Matched-6K Comparison.}
For the cross-method comparison, we
run all Qwen3-VL methods with a strict 6K active-context ceiling in the
no-clarification setting on the full StreamBench and OVO-Bench evaluations.
Each method retains its existing memory design. {}{Compression or eviction precedes most queries for the
persistent-memory methods. Supplementary Sec.~\ref*{sec:appendix-matched-6k}
details the memory settings and pressure.}

\begin{table}[t]

\centering
\caption{Matched-6K Qwen3-VL
results without clarification. Accuracy is reported in percent, TTFT in
seconds, and ProcFPS as online processing throughput.}
\label{tab:matched-6k}
\setlength{\tabcolsep}{1.6pt}
\renewcommand{\arraystretch}{0.95}
\scriptsize
\resizebox{\linewidth}{!}{%
\begin{tabular}{@{}l rrrr rrrr@{}}
\toprule
& \multicolumn{4}{c}{StreamBench} & \multicolumn{4}{c}{OVO-Bench} \\
\cmidrule(lr){2-5}\cmidrule(l){6-9}
Method & Micro & Macro & TTFT$\downarrow$ & ProcFPS$\uparrow$
       & Micro & Macro & TTFT$\downarrow$ & ProcFPS$\uparrow$ \\
\midrule
StreamChat   & 52.8 & 45.4 & 0.314 & 3.42  & 50.5 & 46.2 & 0.198 & 3.68 \\
StreamBridge & 57.7 & 45.7 & 1.170 & 24.51 & 47.7 & 48.8 & 1.197 & 44.86 \\
StreamingVLM & 56.7 & 47.8 & 0.064 & 8.86 & 47.0 & 46.3 & 0.071 & 9.59 \\
Ours         & \bf58.0 & \bf51.3 & 0.232 & 8.46 & \bf56.9 & \bf52.0 & 0.252 & 9.23 \\
\bottomrule
\end{tabular}}
\end{table}

{}{Table~\ref{tab:matched-6k} shows that StreamClarify achieves the highest}
Micro and Macro Accuracy on both benchmarks. Its margins over the strongest
baseline on StreamBench are $+0.3$ Micro points (95\% CI $[-2.3,+3.0]$) and
$+3.4$ Macro points (95\% CI $[+1.2,+5.7]$). On OVO-Bench, the corresponding
margins are $+6.4$ Micro points (95\% CI $[+1.1,+11.8]$) and $+3.2$ Macro
points (95\% CI $[-3.4,+9.2]$). StreamClarify also maintains real-time
processing under this budget{}.

\paragraph{Clarification Retrieval Parameters.}

Figure~\ref{fig:oracle_ablation} studies retrieval depth and clarification
rate under oracle user guidance. Using the same 133/281 triggered OVO-Bench
queries, increasing $K$ from 8 to 32 raises Macro Accuracy from 57.2\% to
58.9\%, while mean TTFT increases from 0.918 to 2.253~s. We retain the intermediate
$K{=}16$ setting. Higher clarification rates similarly
improve Macro Accuracy while reducing ProcFPS.

\begin{figure}[h]
\centering
\includegraphics[width=\linewidth]{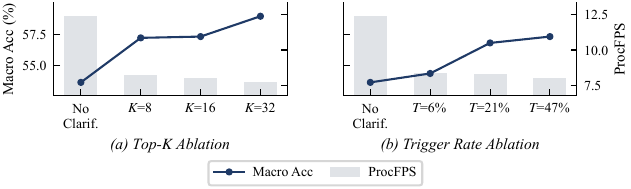}
\caption{Effect of (a) retrieval depth and (b) clarification rate on OVO-Bench
with Qwen3-VL under oracle user guidance. Bars show ProcFPS and lines show Macro Accuracy.}
\label{fig:oracle_ablation}
\end{figure}

\vspace{-0.6em}
\section{Conclusion{}}
StreamClarify studies streaming video QA under bounded active context and
imperfect queries. {} StreamClarify
combines a training-free uncertainty-based clarification trigger with TP-KV
bounded context and clarification-guided archive retrieval.
{}

Several limitations remain. First, the entropy score ranks direct-answer
errors but does not identify whether the underlying cause is ambiguity, missing
context, or reasoning. Second, generated clarification is evaluated in a
single-query setting with a simulated referent-only response. Naturally linked
dialogue, noisy responses, and multi-turn interaction are not covered. Third,
the current benchmarks do not isolate verified old-evidence recovery, so the
memory experiments establish bounded active-context operation rather than a
long-video accuracy advantage. Benchmark construction and answer evaluation
also use LLM-based judgments, which may introduce bias in borderline cases.

Despite these limitations, StreamClarify takes a step toward streaming
assistants that selectively seek focused input before answering.
Future
work will study natural multi-turn clarification and
evaluate memory recovery on long streams with temporally verified supporting
evidence.

\clearpage
\paragraph*{{Acknowledgments.}}
{This work was supported in part by NSF grants CNS-2437204,
CNS-1900875, and CCF-2217144, as well as SBC UT Dallas grant 23011912 and
SBC SUNY Albany grant 3-105218. This research used the Delta advanced
computing and data resource, which is supported by the National Science
Foundation (award OAC-2005572) and the State of Illinois. Delta is a joint
effort of the University of Illinois Urbana-Champaign and its National
Center for Supercomputing Applications. Computing resources were provided
through ACCESS~\cite{boerner2023access} allocation CIS250819. The ACCESS
program is supported by NSF grants 2138259, 2138286, 2138307, 2137603, and
2138296. The opinions expressed in this publication are those of the
authors and do not necessarily reflect those of the funding agencies.}

{\small
\nocite{dao2023flashattention}
\bibliography{main}

\begin{thebibliography}{48}
\providecommand{\natexlab}[1]{#1}
\providecommand{\url}[1]{\texttt{#1}}
\expandafter\ifx\csname urlstyle\endcsname\relax
  \providecommand{\doi}[1]{doi: #1}\else
  \providecommand{\doi}{doi: \begingroup \urlstyle{rm}\Url}\fi

\bibitem[An et~al.(2025)An, Xie, Yang, Zhang, Zhao, Cheng, Wang, Xu, Chen, Zhu,
  et~al.]{an2025llava15}
Xiang An, Yin Xie, Kaicheng Yang, Wenkang Zhang, Xiuwei Zhao, Zheng Cheng,
  Yirui Wang, Songcen Xu, Changrui Chen, Didi Zhu, et~al.
\newblock {LLaVA-OneVision-1.5}: Fully open framework for democratized
  multimodal training.
\newblock \emph{arXiv preprint arXiv:2509.23661}, 2025.

\bibitem[Bai et~al.(2025)Bai, Cai, Chen, Chen, Chen, Cheng, Deng, Ding, Gao,
  Ge, et~al.]{bai2025qwen3}
Shuai Bai, Yuxuan Cai, Ruizhe Chen, Keqin Chen, Xionghui Chen, Zesen Cheng,
  Lianghao Deng, Wei Ding, Chang Gao, Chunjiang Ge, et~al.
\newblock {Qwen3-VL} technical report.
\newblock \emph{arXiv preprint arXiv:2511.21631}, 2025.

\bibitem[Bai et~al.(2024)Bai, Wang, Xiao, He, Han, Zhang, and
  Shou]{bai2024hallucination}
Zechen Bai, Pichao Wang, Tianjun Xiao, Tong He, Zongbo Han, Zheng Zhang, and
  Mike~Zheng Shou.
\newblock Hallucination of multimodal large language models: A survey.
\newblock \emph{arXiv preprint arXiv:2404.18930}, 2024.

\bibitem[Boerner et~al.(2023)Boerner, Deems, Furlani, Knuth, and
  Towns]{boerner2023access}
Timothy~J. Boerner, Stephen Deems, Thomas~R. Furlani, Shelley~L. Knuth, and
  John Towns.
\newblock {ACCESS}: Advancing innovation: {NSF}'s advanced cyberinfrastructure
  coordination ecosystem: Services \& support.
\newblock In \emph{Practice and Experience in Advanced Research Computing},
  pages 173--176. ACM, 2023.

\bibitem[Cai et~al.(2025)Cai, Zhang, Gao, Liu, Li, Liu, Lu, Xiong, Dong, Hu,
  et~al.]{caipyramidkv}
Zefan Cai, Yichi Zhang, Bofei Gao, Yuliang Liu, Yucheng Li, Tianyu Liu, Keming
  Lu, Wayne Xiong, Yue Dong, Junjie Hu, et~al.
\newblock {PyramidKV}: Dynamic {KV} cache compression based on pyramidal
  information funneling.
\newblock In \emph{Second Conference on Language Modeling}, 2025.

\bibitem[Chen et~al.(2024)Chen, Lv, Wu, Lin, Song, Gao, Liu, Gao, Mao, and
  Shou]{chen2024videollm}
Joya Chen, Zhaoyang Lv, Shiwei Wu, Kevin~Qinghong Lin, Chenan Song, Difei Gao,
  Jia-Wei Liu, Ziteng Gao, Dongxing Mao, and Mike~Zheng Shou.
\newblock {VideoLLM-online}: Online video large language model for streaming
  video.
\newblock In \emph{Proceedings of the IEEE/CVF Conference on Computer Vision
  and Pattern Recognition}, pages 18407--18418, 2024.

\bibitem[Chen et~al.(2025{\natexlab{a}})Chen, Tao, Shao, and
  Wang]{chen2025streamingtom}
Xueyi Chen, Keda Tao, Kele Shao, and Huan Wang.
\newblock {StreamingTOM}: Streaming token compression for efficient video
  understanding.
\newblock \emph{arXiv preprint arXiv:2510.18269}, 2025{\natexlab{a}}.

\bibitem[Chen et~al.(2025{\natexlab{b}})Chen, Bai, Wang, Bai, Dai, Lu, and
  Zhang]{chen2025streamkv}
Yilong Chen, Xiang Bai, Zhibin Wang, Chengyu Bai, Yuhan Dai, Ming Lu, and
  Shanghang Zhang.
\newblock {StreamKV}: Streaming video question-answering with segment-based
  {KV} cache retrieval and compression.
\newblock \emph{arXiv preprint arXiv:2511.07278}, 2025{\natexlab{b}}.

\bibitem[Dao(2024)]{dao2023flashattention}
Tri Dao.
\newblock {FlashAttention-2}: Faster attention with better parallelism and work
  partitioning.
\newblock In \emph{International Conference on Learning Representations}, 2024.

\bibitem[Di et~al.(2025)Di, Yu, Zhang, Li, Zhong, Cheng, Li, He, Shu, and
  Jiang]{di2025streaming}
Shangzhe Di, Zhelun Yu, Guanghao Zhang, Haoyuan Li, Tao Zhong, Hao Cheng, Bolin
  Li, Wanggui He, Fangxun Shu, and Hao Jiang.
\newblock Streaming video question-answering with in-context video {KV}-cache
  retrieval.
\newblock In \emph{International Conference on Learning Representations}, 2025.

\bibitem[Ding et~al.(2025)Ding, Wu, Yang, Jiang, Zhang, Bai, Chen, and
  Cao]{ding2025streammind}
Xin Ding, Hao Wu, Yifan Yang, Shiqi Jiang, Qianxi Zhang, Donglin Bai, Zhibo
  Chen, and Ting Cao.
\newblock {StreamMind}: Unlocking full frame rate streaming video dialogue
  through event-gated cognition.
\newblock In \emph{Proceedings of the IEEE/CVF International Conference on
  Computer Vision}, pages 13448--13459, 2025.

\bibitem[Farquhar et~al.(2024)Farquhar, Kossen, Kuhn, and
  Gal]{farquhar2024detecting}
Sebastian Farquhar, Jannik Kossen, Lorenz Kuhn, and Yarin Gal.
\newblock Detecting hallucinations in large language models using semantic
  entropy.
\newblock \emph{Nature}, 630\penalty0 (8017):\penalty0 625--630, 2024.

\bibitem[Fu et~al.(2025)Fu, Dai, Luo, Li, Ren, Zhang, Wang, Zhou, Shen, Zhang,
  et~al.]{fu2025video}
Chaoyou Fu, Yuhan Dai, Yongdong Luo, Lei Li, Shuhuai Ren, Renrui Zhang, Zihan
  Wang, Chenyu Zhou, Yunhang Shen, Mengdan Zhang, et~al.
\newblock {Video-MME}: The first-ever comprehensive evaluation benchmark of
  multi-modal {LLM}s in video analysis.
\newblock In \emph{Proceedings of the IEEE/CVF Conference on Computer Vision
  and Pattern Recognition}, pages 24108--24118, 2025.

\bibitem[{Google DeepMind}(2026)]{google2026gemini36flash}
{Google DeepMind}.
\newblock {Gemini 3.6 Flash Model Card}.
\newblock \url{https://deepmind.google/models/model-cards/gemini-3-6-flash/},
  2026.

\bibitem[He et~al.(2024)He, Zhang, Jin, Yuan, and Yiu]{he2024tubench}
Xingwei He, Qianru Zhang, A-Long Jin, Yuan Yuan, and Siu-Ming Yiu.
\newblock {TUBench}: Benchmarking large vision-language models on
  trustworthiness with unanswerable questions.
\newblock \emph{arXiv preprint arXiv:2410.04107}, 2024.

\bibitem[Jian et~al.(2025)Jian, Yu, Yang, Ren, and Zhang]{jian2025teaching}
Pu Jian, Donglei Yu, Wen Yang, Shuo Ren, and Jiajun Zhang.
\newblock Teaching vision-language models to ask: Resolving ambiguity in visual
  questions.
\newblock In \emph{Proceedings of the 63rd Annual Meeting of the Association
  for Computational Linguistics (Volume 1: Long Papers)}, pages 3619--3638,
  2025.

\bibitem[Kadavath et~al.(2022)Kadavath, Conerly, Askell, Henighan, Drain,
  Perez, Schiefer, Hatfield-Dodds, DasSarma, Tran-Johnson,
  et~al.]{kadavath2022language}
Saurav Kadavath, Tom Conerly, Amanda Askell, Tom Henighan, Dawn Drain, Ethan
  Perez, Nicholas Schiefer, Zac Hatfield-Dodds, Nova DasSarma, Eli
  Tran-Johnson, et~al.
\newblock Language models (mostly) know what they know.
\newblock \emph{arXiv preprint arXiv:2207.05221}, 2022.

\bibitem[Kim et~al.(2025{\natexlab{a}})Kim, Yang, Shin, and Kim]{kim2025v}
Donghyuk Kim, Sejeong Yang, Wonjin Shin, and Joo-Young Kim.
\newblock {V-Rex}: Real-time streaming video {LLM} acceleration via dynamic
  {KV} cache retrieval.
\newblock \emph{arXiv preprint arXiv:2512.12284}, 2025{\natexlab{a}}.

\bibitem[Kim et~al.(2025{\natexlab{b}})Kim, Shim, Choi, and
  Chang]{kim2025infinipot}
Minsoo Kim, Kyuhong Shim, Jungwook Choi, and Simyung Chang.
\newblock {InfiniPot-V}: Memory-constrained {KV} cache compression for
  streaming video understanding.
\newblock In \emph{Advances in Neural Information Processing Systems}, pages
  138983--139013, 2025{\natexlab{b}}.

\bibitem[Kuhn et~al.(2023)Kuhn, Gal, and Farquhar]{kuhn2023semantic}
Lorenz Kuhn, Yarin Gal, and Sebastian Farquhar.
\newblock Semantic uncertainty: Linguistic invariances for uncertainty
  estimation in natural language generation.
\newblock In \emph{International Conference on Learning Representations}, 2023.

\bibitem[Lin et~al.(2024)Lin, Fang, Chen, Wan, Luo, Li, Liu, and
  Sun]{lin2024streamingbench}
Junming Lin, Zheng Fang, Chi Chen, Zihao Wan, Fuwen Luo, Peng Li, Yang Liu, and
  Maosong Sun.
\newblock {StreamingBench}: Assessing the gap for {MLLM}s to achieve streaming
  video understanding.
\newblock \emph{arXiv preprint arXiv:2411.03628}, 2024.

\bibitem[Liu et~al.(2024)Liu, Xue, Chen, Chen, Zhao, Wang, Hou, Li, and
  Peng]{liu2024survey}
Hanchao Liu, Wenyuan Xue, Yifei Chen, Dapeng Chen, Xiutian Zhao, Ke Wang,
  Liping Hou, Rongjun Li, and Wei Peng.
\newblock A survey on hallucination in large vision-language models.
\newblock \emph{arXiv preprint arXiv:2402.00253}, 2024.

\bibitem[Ning et~al.(2025)Ning, Liu, Jin, Li, Ding, Guo, and
  Zhao]{ning2025livevlm}
Zhenyu Ning, Guangda Liu, Qihao Jin, Chengwei Li, Wenchao Ding, Minyi Guo, and
  Jieru Zhao.
\newblock {LiveVLM}: Efficient online video understanding via
  streaming-oriented {KV} cache and retrieval.
\newblock \emph{arXiv preprint arXiv:2505.15269}, 2025.

\bibitem[Niu et~al.(2025)Niu, Li, Miao, Ge, Zhou, He, Dong, Duan, Ding, Qian,
  et~al.]{niu2025ovo}
Junbo Niu, Yifei Li, Ziyang Miao, Chunjiang Ge, Yuanhang Zhou, Qihao He, Xiaoyi
  Dong, Haodong Duan, Shuangrui Ding, Rui Qian, et~al.
\newblock {OVO-Bench}: How far is your {Video-LLM}s from real-world online
  video understanding?
\newblock In \emph{Proceedings of the IEEE/CVF Conference on Computer Vision
  and Pattern Recognition}, pages 18902--18913, 2025.

\bibitem[Patel and Patel(2025)]{patel2025cacheflow}
Shrenik Patel and Daivik Patel.
\newblock {CacheFlow}: Compressive streaming memory for efficient long-form
  video understanding.
\newblock \emph{arXiv preprint arXiv:2511.13644}, 2025.

\bibitem[Qian et~al.(2025)Qian, Ding, Dong, Zhang, Zang, Cao, Lin, and
  Wang]{qian2025dispider}
Rui Qian, Shuangrui Ding, Xiaoyi Dong, Pan Zhang, Yuhang Zang, Yuhang Cao,
  Dahua Lin, and Jiaqi Wang.
\newblock {Dispider}: Enabling video {LLM}s with active real-time interaction
  via disentangled perception, decision, and reaction.
\newblock In \emph{Proceedings of the IEEE/CVF Conference on Computer Vision
  and Pattern Recognition}, pages 24045--24055, 2025.

\bibitem[Singh et~al.(2025)Singh, Fry, Perelman, Tart, Ganesh, El-Kishky,
  McLaughlin, Low, Ostrow, Ananthram, et~al.]{singh2025openai}
Aaditya Singh, Adam Fry, Adam Perelman, Adam Tart, Adi Ganesh, Ahmed El-Kishky,
  Aidan McLaughlin, Aiden Low, AJ Ostrow, Akhila Ananthram, et~al.
\newblock {OpenAI} {GPT-5} system card.
\newblock \emph{arXiv preprint arXiv:2601.03267}, 2025.

\bibitem[Tao et~al.(2025)Tao, Qin, You, Sui, and Wang]{tao2025dycoke}
Keda Tao, Can Qin, Haoxuan You, Yang Sui, and Huan Wang.
\newblock {DyCoke}: Dynamic compression of tokens for fast video large language
  models.
\newblock In \emph{Proceedings of the IEEE/CVF Conference on Computer Vision
  and Pattern Recognition}, pages 18992--19001, 2025.

\bibitem[Testoni and Fern{\'a}ndez(2024)]{testoni2024asking}
Alberto Testoni and Raquel Fern{\'a}ndez.
\newblock Asking the right question at the right time: Human and model
  uncertainty guidance to ask clarification questions.
\newblock In \emph{Proceedings of the 18th Conference of the European Chapter
  of the Association for Computational Linguistics (Volume 1: Long Papers)},
  pages 258--275, 2024.

\bibitem[Tschannen et~al.(2025)Tschannen, Gritsenko, Wang, Naeem,
  Alabdulmohsin, Parthasarathy, Evans, Beyer, Xia, Mustafa,
  et~al.]{tschannen2025siglip}
Michael Tschannen, Alexey Gritsenko, Xiao Wang, Muhammad~Ferjad Naeem, Ibrahim
  Alabdulmohsin, Nikhil Parthasarathy, Talfan Evans, Lucas Beyer, Ye Xia, Basil
  Mustafa, et~al.
\newblock {SigLIP} 2: Multilingual vision-language encoders with improved
  semantic understanding, localization, and dense features.
\newblock \emph{arXiv preprint arXiv:2502.14786}, 2025.

\bibitem[Wang et~al.(2025{\natexlab{a}})Wang, Feng, Lai, Xu, Li, Ge, Dehghan,
  Cao, and Huang]{wang2025streambridge}
Haibo Wang, Bo Feng, Zhengfeng Lai, Mingze Xu, Shiyu Li, Weifeng Ge, Afshin
  Dehghan, Meng Cao, and Ping Huang.
\newblock {StreamBridge}: Turning your offline video large language model into
  a proactive streaming assistant.
\newblock In \emph{Advances in Neural Information Processing Systems}, pages
  132332--132359, 2025{\natexlab{a}}.

\bibitem[Wang et~al.(2025{\natexlab{b}})Wang, Gao, Gu, Pu, Cui, Wei, Liu, Jing,
  Ye, Shao, et~al.]{wang2025internvl3}
Weiyun Wang, Zhangwei Gao, Lixin Gu, Hengjun Pu, Long Cui, Xingguang Wei,
  Zhaoyang Liu, Linglin Jing, Shenglong Ye, Jie Shao, et~al.
\newblock {InternVL3.5}: Advancing open-source multimodal models in
  versatility, reasoning, and efficiency.
\newblock \emph{arXiv preprint arXiv:2508.18265}, 2025{\natexlab{b}}.

\bibitem[Wang et~al.(2025{\natexlab{c}})Wang, He, Hong, Cheng, Zhang, Qi, Ding,
  Gu, Huang, Xu, et~al.]{wang2025lvbench}
Weihan Wang, Zehai He, Wenyi Hong, Yean Cheng, Xiaohan Zhang, Ji Qi, Ming Ding,
  Xiaotao Gu, Shiyu Huang, Bin Xu, et~al.
\newblock {LVBench}: An extreme long video understanding benchmark.
\newblock In \emph{Proceedings of the IEEE/CVF International Conference on
  Computer Vision}, pages 22958--22967, 2025{\natexlab{c}}.

\bibitem[Wang et~al.(2024)Wang, Bingham, Yu, Le, Luong, and
  Ghiasi]{wang2024haloquest}
Zhecan Wang, Garrett Bingham, Adams~Wei Yu, Quoc~V Le, Thang Luong, and Golnaz
  Ghiasi.
\newblock {HaloQuest}: A visual hallucination dataset for advancing multimodal
  reasoning.
\newblock In \emph{European Conference on Computer Vision}, pages 288--304.
  Springer, 2024.

\bibitem[Whitehead et~al.(2022)Whitehead, Petryk, Shakib, Gonzalez, Darrell,
  Rohrbach, and Rohrbach]{whitehead2022reliable}
Spencer Whitehead, Suzanne Petryk, Vedaad Shakib, Joseph Gonzalez, Trevor
  Darrell, Anna Rohrbach, and Marcus Rohrbach.
\newblock Reliable visual question answering: Abstain rather than answer
  incorrectly.
\newblock In \emph{European Conference on Computer Vision}, pages 148--166.
  Springer, 2022.

\bibitem[Wu et~al.(2024)Wu, Li, Chen, and Li]{wu2024longvideobench}
Haoning Wu, Dongxu Li, Bei Chen, and Junnan Li.
\newblock {LongVideoBench}: A benchmark for long-context interleaved
  video-language understanding.
\newblock In \emph{Advances in Neural Information Processing Systems}, pages
  28828--28857, 2024.

\bibitem[Xiao et~al.(2024)Xiao, Tian, Chen, Han, and Lewis]{xiao2023efficient}
Guangxuan Xiao, Yuandong Tian, Beidi Chen, Song Han, and Mike Lewis.
\newblock Efficient streaming language models with attention sinks.
\newblock In \emph{International Conference on Learning Representations}, 2024.

\bibitem[Xiong et~al.(2025)Xiong, Yang, Yu, Zhuge, Zhang, Zhu, and
  Lu]{xiong2025streaming}
Haomiao Xiong, Zongxin Yang, Jiazuo Yu, Yunzhi Zhuge, Lu Zhang, Jiawen Zhu, and
  Huchuan Lu.
\newblock Streaming video understanding and multi-round interaction with
  memory-enhanced knowledge.
\newblock In \emph{International Conference on Learning Representations}, 2025.

\bibitem[Xu et~al.(2025)Xu, Xiao, Chen, He, Peng, Lu, and
  Han]{xu2025streamingvlm}
Ruyi Xu, Guangxuan Xiao, Yukang Chen, Liuning He, Kelly Peng, Yao Lu, and Song
  Han.
\newblock {StreamingVLM}: Real-time understanding for infinite video streams.
\newblock \emph{arXiv preprint arXiv:2510.09608v1}, 2025.

\bibitem[Yang et~al.(2025{\natexlab{a}})Yang, Li, Yang, Zhang, Hui, Zheng, Yu,
  Gao, Huang, Lv, et~al.]{yang2025qwen3}
An Yang, Anfeng Li, Baosong Yang, Beichen Zhang, Binyuan Hui, Bo Zheng, Bowen
  Yu, Chang Gao, Chengen Huang, Chenxu Lv, et~al.
\newblock {Qwen3} technical report.
\newblock \emph{arXiv preprint arXiv:2505.09388}, 2025{\natexlab{a}}.

\bibitem[Yang et~al.(2024)Yang, Han, Gao, Hu, Zhang, and
  Zhao]{yang2024pyramidinfer}
Dongjie Yang, Xiaodong Han, Yan Gao, Yao Hu, Shilin Zhang, and Hai Zhao.
\newblock {PyramidInfer}: Pyramid {KV} cache compression for high-throughput
  {LLM} inference.
\newblock In \emph{Findings of the Association for Computational Linguistics:
  ACL 2024}, pages 3258--3270, 2024.

\bibitem[Yang et~al.(2025{\natexlab{b}})Yang, Tang, Zhao, Zhuang, Lu, An, Hu,
  Zhang, Swikir, He, Ge, Khan, and Razzak]{yang2025streamagent}
Haolin Yang, Feilong Tang, Lingxiao Zhao, Xinlin Zhuang, Yifan Lu, Xiang An,
  Ming Hu, Xiaofeng Zhang, Abdalla Swikir, Junjun He, Zongyuan Ge,
  Muhammad~Haris Khan, and Imran Razzak.
\newblock {StreamAgent}: Towards anticipatory agents for streaming video
  understanding.
\newblock \emph{arXiv preprint arXiv:2508.01875}, 2025{\natexlab{b}}.

\bibitem[Yang et~al.(2025{\natexlab{c}})Yang, Zhang, Hu, Wang, Qian, Wen, Yang,
  Gao, Dong, and Xu]{yang2025livestar}
Zhenyu Yang, Kairui Zhang, Yuhang Hu, Bing Wang, Shengsheng Qian, Bin Wen, Fan
  Yang, Tingting Gao, Weiming Dong, and Changsheng Xu.
\newblock {LiveStar}: Live streaming assistant for real-world online video
  understanding.
\newblock In \emph{Advances in Neural Information Processing Systems}, pages
  31266--31304, 2025{\natexlab{c}}.

\bibitem[Yao et~al.(2025)Yao, Li, Wei, Li, Ren, Liu, Ouyang, Wang, Li, Li,
  et~al.]{yao2025timechat}
Linli Yao, Yicheng Li, Yuancheng Wei, Lei Li, Shuhuai Ren, Yuanxin Liu, Kun
  Ouyang, Lean Wang, Shicheng Li, Sida Li, et~al.
\newblock {TimeChat-Online}: 80\% visual tokens are naturally redundant in
  streaming videos.
\newblock In \emph{Proceedings of the 33rd ACM International Conference on
  Multimedia}, pages 10807--10816, 2025.

\bibitem[Yu et~al.(2025)Yu, Wang, Wang, Huang, Ma, He, Cai, Chen, Huang, Zhao,
  et~al.]{yu2025minicpm}
Tianyu Yu, Zefan Wang, Chongyi Wang, Fuwei Huang, Wenshuo Ma, Zhihui He,
  Tianchi Cai, Weize Chen, Yuxiang Huang, Yuanqian Zhao, et~al.
\newblock {MiniCPM-V 4.5}: Cooking efficient {MLLM}s via architecture, data,
  and training recipe.
\newblock \emph{arXiv preprint arXiv:2509.18154}, 2025.

\bibitem[Zhang et~al.(2025{\natexlab{a}})Zhang, Wang, Tang, Liu, Feng, and
  Jin]{zhang2025flash}
Haoji Zhang, Yiqin Wang, Yansong Tang, Yong Liu, Jiashi Feng, and Xiaojie Jin.
\newblock {Flash-VStream}: Efficient real-time understanding for long video
  streams.
\newblock In \emph{Proceedings of the IEEE/CVF International Conference on
  Computer Vision}, pages 21059--21069, 2025{\natexlab{a}}.

\bibitem[Zhang et~al.(2025{\natexlab{b}})Zhang, Knox, and Choi]{zhangmodeling}
Michael~JQ Zhang, W~Bradley Knox, and Eunsol Choi.
\newblock Modeling future conversation turns to teach {LLMs} to ask clarifying
  questions.
\newblock In \emph{International Conference on Learning Representations},
  2025{\natexlab{b}}.

\bibitem[Zhang et~al.(2026)Zhang, Shi, and Yang]{zhang2026weavetime}
Yulin Zhang, Cheng Shi, and Sibei Yang.
\newblock {WeaveTime}: Stream from earlier frames into emergent memory in
  {VideoLLMs}.
\newblock \emph{arXiv preprint arXiv:2602.22142}, 2026.

\end{thebibliography}
}

\fi

\ifdefined\MainOnly
\else
\ifdefined\SupplementOnly
\else
\clearpage
\raggedbottom
\appendix
\fi

\section{\texorpdfstring{Notation}{Notation}}
\label{sec:appendix-notation}

\begin{center}
\captionsetup{type=table,hypcap=false}
\caption{Notation used by StreamClarify.}
\label{tab:notation}
\setlength{\tabcolsep}{2pt}
\renewcommand{\arraystretch}{0.9}
\scriptsize
\begin{tabular}{@{}p{0.27\linewidth}p{0.68\linewidth}@{}}
\toprule
Symbol & Meaning \\
\midrule
$\mathcal{V}=\{F_1,F_2,\dots\}$ & Continuous input stream, with frame $F_t$ at time $t$ \\
$\mathcal{V}_{\le t_i}$ & Frames observed up to query time $t_i$ \\
$\langle V_i,Q_i,A_i\rangle$ & Round-$i$ video context, user query, and model answer \\
$\langle R_i\rangle$ & User-side guidance after clarification in round $i$ \\
$\mathcal{M}$ & Bounded active-context budget \\
$L_{\text{max}}$ & Maximum active-context token budget for the KV cache \\
$L_{\text{comp}}$ & Maximum pyramid/compression level \\
$\ell$ & Pyramid level of a token, where $\ell=0$ is full resolution \\
$2^{\ell}$ & Subsampling stride at pyramid level $\ell$ \\
$\rho$ & Compression ratio controlling retained historical tokens \\
$\mathcal{KV}$ & Active KV cache produced by backbone prefilling \\
$|\mathcal{KV}|$ & Active KV cache size, constrained by $L_{\text{max}}$ \\
$\mathcal{T}_{\text{rec}}$ & Recency-protection tokens kept at full resolution \\
$\mathcal{T}_{\text{hist}}$ & Historical tokens eligible for compression or eviction \\
$\mathcal{I}_{\text{vis}}$ & Backbone vision encoder for answer generation \\
$\mathbf{v}_t$ & Backbone vision embeddings for frame $t$ \\
$\mathcal{I}_{\text{ret}}$ & Retrieval-key encoder for semantic search, instantiated with SigLIP2 \\
$\mathbf{e}_t$ & Image-text retrieval key for frame $t$ \\
$\mathcal{A}$ & CPU-side archive of $(\mathbf{v}_t,\mathbf{e}_t,t)$ entries stored separately from the active KV cache \\
$p$ & Probability assigned to \texttt{[Yes]} in the Stage~1 judgment \\
$H$ & Binary entropy of the Stage~1 \texttt{[Yes]/[No]} judgment \\
$\tau$ & Entropy threshold for triggering clarification \\
$K$ & Number of top retrieved clips used after clarification \\
$\mathcal{T}^{*}$ & Historical and recent tokens retained in the updated TP-KV context \\
\bottomrule
\end{tabular}
\end{center}

\section{\texorpdfstring{Memory Management Algorithm}{Memory Management Algorithm}}
\label{sec:appendix-memory-algorithm}

\begingroup
\setlength{\intextsep}{0pt}
\begin{algorithm}[h]
\footnotesize
\caption{Online TP-KV Update and Archive Construction}
\label{alg:memory}
\SetKwInOut{Input}{Inputs}
\SetKwInOut{Init}{Initializations}
\SetKwComment{Comment}{\textit{// }}{}
\Input{frames $[F_1, F_2, \ldots, F_t, \ldots]$}
\Init{$\mathcal{KV}\!=\![\cdot]$: active context;\
$\mathcal{A}\!=\![\cdot]$: retrieval memory;\
$L_{\text{max}}$: max tokens;\ $\rho$: compression ratio;\
$L_{\text{comp}}$: max compression level}
\While{$F_t$ arrives}{
    $\mathbf{v}_t \leftarrow \mathcal{I}_{\text{vis}}(F_t)$;\
    $\mathcal{KV} \leftarrow \text{Prefill}(\mathcal{KV}, \mathbf{v}_t)$;\
    $\mathbf{e}_t \leftarrow \mathcal{I}_{\text{ret}}(F_t)$;\
    $\mathcal{A}.\text{append}((\mathbf{v}_t, \mathbf{e}_t, t))$
    \Comment*[r]{encode \& archive}
    \If{$|\mathcal{KV}| > L_{\text{max}}$}{
        $\mathcal{T}_{\text{rec}} \leftarrow$ tokens within recent window;\
        $\mathcal{T}_{\text{hist}} \leftarrow$ remaining tokens\;
        \For{each $\mathbf{v} \in \mathcal{T}_{\textup{hist}}$ where
        $\textup{level}(\mathbf{v}) < L_{\textup{comp}}$}{
            compress $\mathbf{v}$ via stride sampling;\
            $\text{level}(\mathbf{v}) \mathrel{+}= 1$
            \Comment*[r]{promote to next pyramid level}
        }
        \If{$|\mathcal{T}_{\textup{hist}}| >
        (L_{\text{max}} - |\mathcal{T}_{\textup{rec}}|) \cdot \rho$}{
            evict $\mathbf{v}$ where $\text{level}(\mathbf{v}) \geq
            L_{\text{comp}}$, oldest first
            \Comment*[r]{ratio unsatisfied: evict max-level tokens}
        }
        $\mathcal{T}^* \leftarrow \mathcal{T}_{\text{hist}} \oplus
        \mathcal{T}_{\text{rec}}$;\ order $\mathcal{T}^*$ chronologically\;
        $\mathcal{KV} \leftarrow \text{Prefill}(\mathcal{T}^*)$
        \Comment*[r]{rebuild KV in single forward pass}
    }
    $t \mathrel{+}= 1$\;
}
\end{algorithm}
\endgroup

\section{\texorpdfstring{{}{Benchmark Construction and Validation}}{Benchmark Construction and Validation}}
\label{sec:appendix-uq-audit}

{For each source video, MiniCPM-V 4.5~\cite{yu2025minicpm} generates
dense captions at 2~FPS and 448-pixel resolution from overlapping 6-second
windows with a 3-second stride. The caption prompt includes the source
questions but not their answers, and the resulting captions are aligned to
each query timestamp.}
Timestamp-aligned dense captions are available for all 1,556
StreamBench queries.

For Referential Ambiguity QA, timestamp-aligned captions identify contexts with
multiple plausible referents and guide the query rewrite. Human review covers
all 404 candidates and removes insufficiently ambiguous rewrites and cases whose
referent remains recoverable from the original annotations, leaving 302 pairs.

For StreamBench Unanswerable QA, the construction run generates 1,563
distractor sets from 995 source pairs, including retries. Caption-based
validation rejects 826 sets in which at least one distractor is supported before
the query timestamp and accepts 737, each associated with a unique source QA.
Balanced split assembly retains 389 of these validated candidates.

Raw-video validation covers all 389 retained StreamBench {Unanswerable QA }instances. The 151
OVO-Bench HLD instances are inherited without modification and are not part of
this validation. {}{For each
StreamBench instance, we clip the video to $[0,t_q]$, sample at 1~FPS, and
remove audio.} The resulting
input contains no frame after the query timestamp.

Gemini 3.6 Flash performs three successive screening rounds, beginning
with all 389 visual prefixes. An instance remains flagged only if the available
prefix shows that the object or event assumed by the question is present and
directly supports one particular answer option other than the abstention option.
Both conditions are required so that a partial visual match is not counted as a
label error. Seven instances remain flagged after all three rounds and are
inspected by a human reviewer using the original visual prefixes. {}{Human review
confirms the original Unanswerable label for all seven reviewed instances.} Thus, the validation
finds no human-confirmed label error among the 389 screened instances. Gemini is
used only to identify candidates, while human review provides the final
decision.

\section{\texorpdfstring{{}{Experimental Setup}}{Experimental Setup}}
\label{sec:appendix-implementation}

\subsection{\texorpdfstring{{Evaluation Scope and Baselines}}{Evaluation Scope and Baselines}}

\begin{table*}[t]
\centering
\caption{Dataset, backbone, and clarification scope of the main experiments
and ablations.}
\label{tab:evaluation-scope}
\scriptsize
\setlength{\tabcolsep}{4pt}
\renewcommand{\arraystretch}{1.02}
\begin{tabular}{@{}p{0.20\linewidth}p{0.32\linewidth}p{0.20\linewidth}p{0.20\linewidth}@{}}
\toprule
Analysis & Evaluation population & Backbone & Clarification setting \\
\midrule
Main comparison & StreamBench ($1{,}556$ QA) and OVO-Bench ($281$ QA), reported separately
& LLaVA-OV-1.5, InternVL-3.5, Qwen3-VL & Disabled \\
\addlinespace[3pt]
Oracle-guided analysis & StreamBench and OVO-Bench pooled ($1{,}837$ QA)
& LLaVA-OV-1.5, InternVL-3.5, Qwen3-VL & Oracle guidance after triggering \\
\addlinespace[3pt]
Generated-question analysis & Ambiguous QA from StreamBench and OVO-Bench, pooled ($302$ QA)
& Qwen3-VL & No
clarification, Generated Clarification ({random-gated, }entropy-gated{,} and all-query), Direct
Retrieval, and Oracle Guidance \\
\addlinespace[3pt]
Pre-answer scoring & OVO-Bench ($281$ QA) & Qwen3-VL & Disabled \\
\addlinespace[3pt]
TP-KV hyperparameters & OVO-Bench ($281$ QA) & Qwen3-VL & Disabled \\
\addlinespace[3pt]
Matched-6K comparison & StreamBench ($1{,}556$ QA) and OVO-Bench ($281$ QA), reported separately
& Qwen3-VL & Disabled \\
\addlinespace[3pt]
Retrieval parameters & OVO-Bench ($281$ QA) & Qwen3-VL & Oracle guidance \\
\addlinespace[3pt]
\multirow{2}{*}{Prompt perturbations}
& OVO-Bench ($281$ QA) & Qwen3-VL & Disabled \\
& StreamBench ($1{,}556$ QA) & Qwen3-VL, InternVL-3.5 & Disabled \\
\bottomrule
\end{tabular}
\end{table*}

{}{The evaluated splits contain 1,556 questions from 260 StreamBench
videos and 281 questions from 95 OVO-Bench videos.}
{}{The StreamBench split comprises 913 Standard, 254 Ambiguous, and
389 Unanswerable questions. The corresponding OVO-Bench counts are
82 Standard, 48 Ambiguous, and 151 Unanswerable questions.}
Table~\ref{tab:evaluation-scope} distinguishes experiments that report the two
benchmarks separately from those that pool their QA pairs. Experiments use the
corresponding main configuration unless the evaluated factor is explicitly
varied.

\paragraph{Baseline Protocol.}
All streaming methods are evaluated at the same annotated query timestamps using
the same QA prompting protocol, answer judge, and evaluation pipeline.
Training-based methods use their officially released weights. For the
backbone-matched training-free comparison, StreamChat~\cite{xiong2025streaming},
StreamBridge~\cite{wang2025streambridge}, StreamingVLM~\cite{xu2025streamingvlm},
{}{and StreamClarify use the same three frozen backbones without}
task-specific fine-tuning. This controls for differences in backbone capability. {StreamingVLM uses its
inference-time cache policy on these frozen backbones. Its original supervised
fine-tuning procedure is not reproduced.}

{}{GPT-5-mini~\cite{singh2025openai}
is evaluated through the OpenAI API. For each query, up to 64 frames are sampled
at 1~FPS from the recent video window ending at the query timestamp and
submitted as a single batch.}

\subsection{\texorpdfstring{{Evaluation Protocol}}{Evaluation Protocol}}

Answer correctness is determined by an LLM-based semantic judge
(Qwen3-14B)~\cite{yang2025qwen3} using task-specific prompts conditioned on dense
offline captions as scene-level reference context. General knowledge questions
are judged for factual correctness, while all other questions are judged for
semantic equivalence against the ground-truth answer. Ambiguous questions also
require entity-level consistency. Human spot-checks on 50 sampled videos are
conducted to verify and improve judgment quality.

Micro Accuracy is computed as the overall accuracy across all QA pairs. Macro
Accuracy is the unweighted mean of per-scenario accuracies across Standard QA,
{}{Referential Ambiguity QA, and Unanswerable QA. We measure system efficiency using}
time-to-first-token (TTFT, seconds) and online processing throughput (ProcFPS), measured
in frames per second.

For trigger diagnostics, incorrect
initial answers are treated as positives. For each backbone, Random and
Entropy selections {in the pooled oracle-guided comparison }are evaluated against the corresponding no-clarification
{}{answers on the same $1{,}837$ QA pairs.} Trigger rate, Precision, and
Recall describe selected operating points, while Error AUROC measures
threshold-independent ranking using the same labels. We report wrong-to-correct
and correct-to-wrong transitions, together with the resulting change in
accuracy, to measure the effect of clarification.

\subsection{\texorpdfstring{{Backbone and Retrieval Settings}}{Backbone and Retrieval Settings}}

{}{StreamClarify uses three instruction-tuned 8B VLM backbones:}
LLaVA-OV-1.5~{}{\cite{an2025llava15}}, InternVL-3.5~\cite{wang2025internvl3}, and
Qwen3-VL~\cite{bai2025qwen3}. All models are evaluated without task-specific training.
All backbones use bfloat16 precision with
FlashAttention-2{~\cite{dao2023flashattention}}. Input frames are resized to
$448\times448$. Qwen3-VL is sampled at 2~FPS with temporal patch size 2,
yielding 196 vision tokens per frame pair. LLaVA-OV-1.5 and InternVL-3.5 are run
at 1~FPS with 256 tokens per frame, which keeps the active memory usage
comparable across backbones.
Decoding is greedy with a
maximum answer length of 32 tokens. Each open-model inference job uses one NVIDIA A40 GPU.

The main
comparison follows each method's standard configuration. A complementary
Qwen3-VL comparison applies a common 6K active-context budget.
Unless otherwise specified, TP-KV uses a 10K-token budget, compression ratio 0.25, and a 20~s recency window.
For diagnostic clarification experiments, the per-backbone trigger
threshold is set from the no-clarification entropy distribution to obtain the
reported operating trigger rate, and retrieval uses top-$K{=}16$ clips.
{Section~\ref{sec:appendix-matched-6k} details the per-method memory settings and memory pressure in the matched-6K comparison.}

The retrieval-key encoder
{}{$\mathcal{I}_{\text{ret}}$ uses the same fixed}
SigLIP2{~\cite{tschannen2025siglip}} (so400M, patch size~14, resolution~384)
across all backbones. Retrieved top-$K$ clips are prefilled and combined with
the current TP-KV context for {}{answering} after clarification.
Each
backbone uses its native positional encoding when the retrieved clips are
combined with the current TP-KV context.

\section{\texorpdfstring{{}{Additional Experimental Analyses}}{Additional Experimental Analyses}}

\subsection{\texorpdfstring{{}{Paired Confidence Intervals}}{Paired Confidence Intervals}}
\label{sec:appendix-main-ci}

\begin{table}[t]

\centering
\caption{Video-level paired Macro {}{Accuracy} intervals for the
main-table configurations. Differences are StreamClarify minus the
selected baseline and follow the one-decimal accuracies reported in
Table~\ref*{tab:main-comparison}. Differences and interval endpoints are
percentage points. Intervals are computed from paired question-level
correctness and
100,000 whole-video bootstrap replicates.}
\label{tab:main-macro-ci}
\setlength{\tabcolsep}{2.5pt}
\renewcommand{\arraystretch}{0.95}
\scriptsize
\resizebox{\linewidth}{!}{%
\begin{tabular}{@{}llllr@{}}
\toprule
Dataset & Backbone & Selected baseline & Difference & Paired 95\% CI \\
\midrule
StreamBench & LLaVA-OV-1.5 & StreamChat & $+0.2$ & $[-2.4,+2.7]$ \\
StreamBench & InternVL-3.5 & StreamBridge & $+1.1$ & $[-1.4,+3.6]$ \\
StreamBench & Qwen3-VL & StreamingVLM & $+3.3$ & $[+1.0,+5.7]$ \\
OVO-Bench & LLaVA-OV-1.5 & StreamBridge & $+8.2$ & $[+2.2,+13.8]$ \\
OVO-Bench & InternVL-3.5 & StreamBridge & $+5.7$ & $[-1.8,+13.2]$ \\
OVO-Bench & Qwen3-VL & StreamBridge & $+7.8$ & $[+2.4,+12.9]$ \\
\bottomrule
\end{tabular}}
\end{table}

{We estimate uncertainty in the main-table Macro Accuracy differences
using paired whole-video resampling (Table~\ref{tab:main-macro-ci}).}
For each dataset and backbone, we select the strongest training-free
baseline using unrounded Macro Accuracy and hold that selection fixed during
resampling. We then resample complete videos with replacement and recompute
Standard, Ambiguous, and Unanswerable QA accuracy from all sampled questions.
Macro is their unweighted mean.
StreamClarify has higher Macro Accuracy in all six comparisons. The paired
interval excludes zero for StreamBench/Qwen3-VL, OVO-Bench/LLaVA-OV-1.5, and
OVO-Bench/Qwen3-VL.

\subsection{\texorpdfstring{{}{Random-Gated Clarification Control}}{Random-Gated Clarification Control}}
\label{sec:appendix-random-generated}
{We compare entropy gating with uniform random selection of the same
number of queries. The random-gated control in Table~\ref*{tab:generated-clarification}
reuses the same per-query answers and Generated Clarification outcomes,
including the gate and fallback decisions. Each of $1{,}000$ trials uniformly
selects 162 of the 302 queries without replacement. Selected queries use
their saved Generated Clarification outcomes, and unselected queries retain
their no-clarification answers. Mean accuracy is 29.09\%,
matching the theoretical expectation under uniform selection. The trials average 29.52 corrections
and 9.68 new errors. The standard deviation of accuracy across trials is
1.43 percentage points.}

\subsection{\texorpdfstring{Evaluator Sensitivity Analysis}{Evaluator Sensitivity Analysis}}
\label{sec:appendix-evaluator-sensitivity}

\begin{table}[t]

\centering
\caption{Evaluator sensitivity in the generated-question
analysis. Accuracies are percentages. Gains and interval endpoints are
percentage points. Gain compares the
entropy-gated condition with w/o Clarification. Intervals use paired
whole-video resampling.}
\label{tab:cross-model-evaluator}
\setlength{\tabcolsep}{2.0pt}
\renewcommand{\arraystretch}{0.95}
\scriptsize
\resizebox{\linewidth}{!}{%
\begin{tabular}{@{}lrrrrrl@{}}
\toprule
Judge & w/o & Direct & Gen. all & Gen. gated & Oracle & Gated gain (95\% CI) \\
\midrule
Qwen3-14B & 22.5 & 22.8 & 34.8 & 30.8 & 38.4 & $+8.3\ [4.2,12.5]$ \\
GPT-5 & 20.5 & 22.2 & 37.4 & 31.1 & 42.4 & $+10.6\ [6.5,14.9]$ \\
Gemini 3.6 Flash & 29.1 & 26.5 & 43.4 & 37.1 & 50.7 & $+8.0\ [3.3,12.8]$ \\
\bottomrule
\end{tabular}}
\end{table}

Table~\ref*{tab:generated-clarification} uses Qwen3-14B as the primary
semantic answer judge. We assess sensitivity to this choice by asking blinded
GPT-5 and Gemini 3.6 Flash answer judges to score the same answers from w/o
Clarification, Direct Retrieval, Generated Clarification, and Oracle Guidance. Each
judge receives only the question type, question, reference answer, candidate
answer, and query-time caption context. It evaluates the candidate answer
without knowing which experimental condition produced it or how Qwen3-14B
scored it. For the
entropy-gated result, we follow the same selection used in
Table~\ref*{tab:generated-clarification}, taking the Generated Clarification
answer for the 162 selected queries and the w/o Clarification answer for the
remaining 140.
Pairwise raw agreement ranges from 83.2\% to 93.0\%, with Cohen's
$\kappa$ ranging from 0.626 to 0.845. {}{All accuracies are computed over the same 302
questions.} {Table~\ref{tab:cross-model-evaluator} shows that }the
absolute accuracies vary across judges, but the entropy-gated gain remains
positive in all three evaluations.

\subsection{\texorpdfstring{{}{Trigger Score Comparison and Robustness}}{Trigger Score Comparison and Robustness}}
\label{sec:appendix-trigger-robustness}

\paragraph{{}{Directional Score Comparison.}}
We compare entropy with $p(\mathrm{No})$ using the
\texttt{[Yes]/[No]} judgment probabilities on all 1,556 StreamBench queries. For a
normalized binary distribution, $p(\mathrm{No})$, $1-p(\mathrm{Yes})$, and the
signed \texttt{No}-versus-\texttt{Yes} logit margin produce the same ranking.
Table~\ref{tab:entropy-pno} evaluates both rankings against direct-answer
correctness. {These scores come from the oracle-guided run and are
evaluated against the no-clarification error labels.}
At a common 30.5\% selection rate, entropy selects 280 incorrect
answers, compared with 154 for $p(\mathrm{No})$.

\begin{center}

\captionof{table}{Directional score comparison on StreamBench with
Qwen3-VL. Precision and Recall are measured at a common 30.5\% selection rate
($474/1{,}556$ queries).}
\label{tab:entropy-pno}
\scriptsize
\setlength{\tabcolsep}{8pt}
\renewcommand{\arraystretch}{0.95}
\begin{tabular}{@{}lccc@{}}
\toprule
Score & Error AUROC & Precision (\%) & Recall (\%) \\
\midrule
Entropy & \textbf{0.674} & \textbf{59.1} & \textbf{44.4} \\
$p(\mathrm{No})$ & 0.527 & 32.5 & 24.4 \\
\bottomrule
\end{tabular}
\end{center}

The {}{per-category} AUROC values for Standard, Ambiguous, and
Unanswerable QA are $0.592$, $0.571$, and $0.795$ for entropy and $0.676$,
$0.632$, and $0.138$ for
$p(\mathrm{No})$, respectively. Thus, the overall entropy advantage is
concentrated on Unanswerable QA, while the directional score {}{has higher AUROC} on Standard and Ambiguous QA. A possible explanation is that, on
Unanswerable QA, a confident \texttt{[No]} judgment may accompany correct
abstention rather than an error. Entropy does not assign risk to either
direction and instead reflects uncertainty between the two judgments.

\paragraph{{}{Prompt Perturbations.}}
We denote the original prompt
for each evaluated backbone as P0 and vary one
component at a time. P1 replaces \texttt{Output} with \texttt{Respond} and changes one synonym
in the final instruction, while retaining the label definitions and demonstrations.
P2 only reverses the two demonstrations within each \texttt{[Yes]} and
\texttt{[No]} section. P3 replaces
them with two visually grounded and two general-knowledge demonstrations that
do not occur verbatim in the benchmark annotations.
Each prompt is evaluated independently from the beginning of every
video. All other model, decoding, streaming, memory, and evaluation settings
remain fixed. P0 {in Table~\ref{tab:prompt-robustness} }reproduces the original no-clarification outputs and scores.

\begin{center}

\begin{minipage}{\linewidth}
\centering
\captionof{table}{Error AUROC under Stage-1 prompt perturbations.}
\label{tab:prompt-robustness}
\scriptsize
\setlength{\tabcolsep}{5.5pt}
\renewcommand{\arraystretch}{0.95}
\begin{tabular}{@{}lcccc@{}}
\toprule
Dataset / Backbone & P0 & P1 & P2 & P3 \\
\midrule
OVO-Bench / Qwen3-VL & 0.704 & 0.707 & 0.709 & 0.721 \\
StreamBench / Qwen3-VL & 0.675 & 0.667 & 0.663 & 0.680 \\
StreamBench / InternVL & 0.641 & 0.638 & 0.635 & 0.639 \\
\bottomrule
\end{tabular}
\end{minipage}
\end{center}

Table~\ref{tab:prompt-robustness} evaluates each score ranking against
the errors produced by the corresponding prompt's regenerated answers. Across the perturbations,
Error AUROC changes by $-0.011$ to $+0.017$ relative to P0, and Spearman
correlation with the P0 ranking is $0.941$--$0.978$. In a separate analysis, we
compare every prompt's entropy ranking against the same answer errors from P0.
The maximum absolute AUROC change is $0.011$, so the ranking result is not
explained by changes in the Stage-2 answers.
{}{With each backbone's
threshold held fixed, prompt changes alter the trigger rate by at most 4.7
percentage points.} These results measure
ranking and operating-point stability rather than probability calibration.

\section{\texorpdfstring{{}{Memory and Runtime Analysis}}{Memory and Runtime Analysis}}

\subsection{\texorpdfstring{{Matched-6K Memory Analysis}}{Matched-6K Memory Analysis}}
\label{sec:appendix-matched-6k}
The 6K comparison matches active visual context rather than total
storage, and all methods remain within this limit.
{Table~\ref{tab:qwen-memory-configs} lists the per-method memory configurations.}
StreamChat has no persistent
visual KV cache, so the limit applies to the retrieved visual context formed at
query time. Its searchable CPU archive remains available and occupies
292.0~MB per video on average and 565.6~MB at peak. In this no-clarification
comparison, StreamBridge, StreamingVLM, and StreamClarify do not use an external
retrieval archive when answering.

\begin{table*}[t]

\centering
\caption{Qwen3-VL memory settings in the main comparison and
the matched-6K comparison. Active context denotes the visual context available
to the backbone for answering. The 6K limit does not include CPU archive
storage.}
\label{tab:qwen-memory-configs}
\setlength{\tabcolsep}{4pt}
\renewcommand{\arraystretch}{1.05}
\scriptsize
\begin{tabular}{@{}p{0.11\linewidth}p{0.39\linewidth}p{0.40\linewidth}@{}}
\toprule
Method & Main-comparison setting & Matched-6K setting \\
\midrule
StreamChat & No persistent visual KV cache. At each query, its native retrieval
selects frames from a CPU-side archive to form a fresh visual context. & The
archive and retrieval procedure are unchanged, while the retrieved visual
context supplied to the backbone is limited to 6K. \\
StreamBridge & Visual embeddings are accumulated in a memory buffer and
progressively compressed. When a question arrives, the current buffer is
used to prefill the backbone under its native 16K context limit. & The same
query-time prefilling procedure, with the embedding buffer compressed to fit
the 6K active-context limit. \\
StreamingVLM & A sliding window retaining 512 attention-sink tokens and up to
6,192 recent visual tokens, for 6.7K tokens in total. & The same sliding-window
design with 512 attention-sink tokens and up to 5,632 recent visual tokens,
for 6K tokens in total. \\
StreamClarify & TP-KV with a 10K-token budget, a 20~s full-granularity recent
window, and {}{compression ratio $\rho=0.25$}. &
The same TP-KV policy under a 6K budget, preserving the 20~s recent
window and {}{the compression ratio $\rho=0.25$}. \\
\bottomrule
\end{tabular}
\end{table*}

\paragraph{Memory Pressure.}
{We measure how often memory management occurs to verify that the
matched-6K comparison evaluates methods under active memory constraints.}
At the evaluated frame rate, 6K corresponds to about 31~s of dense
visual history{, compared with average stream durations of 253~s and
383~s for StreamBench and OVO-Bench, respectively}. Across the full evaluation,
StreamBridge, StreamingVLM, and StreamClarify apply compression or eviction
in all 355 videos, before 84.8\%, 86.7\%, and 85.2\% of their queries,
respectively. The evaluated StreamClarify 10K and strict-6K
configurations produce 2.14 and 6.19 compression operations per minute of
streamed video. {Under strict 6K, 85.2\% of StreamClarify queries arrive
after compression, compared with 76.7\% under the 10K configuration. The
matched-6K setting therefore evaluates most queries after active-context
management has already occurred.}

\subsection{\texorpdfstring{{Retrieval Archive Growth}}{Retrieval Archive Growth}}
\label{sec:appendix-memory-scalability}

{}{We
measure how the} {}{CPU-side}
retrieval archive grows with streaming duration. {}{We use}
{}{an approximately 30-minute OVO-Bench video and Qwen3-VL-8B-Instruct
with the main experiment's 10K TP-KV configuration.}
The archive maintains
lightweight SigLIP2 retrieval keys and stores Qwen
visual
representations needed to reconstruct retrieved clips. Raw frames are excluded,
and TP-KV is measured separately as the bounded active context rather than as
part of the long-term archive.

The archive size grows monotonically and approximately linearly with video time.
Over the approximately 30-minute stream, the archive stores about 1.8K Qwen
vision clips and indexes about 3.6K frames with SigLIP2 retrieval keys. The Qwen
vision-embedding archive occupies about 2.7~GiB, while the SigLIP2 retrieval index
adds only about 8.5~MiB. Overall, the {}{CPU-side} archive grows at roughly
92~MiB per minute. This growth affects only the searchable CPU-side archive. The
online TP-KV remains bounded by its compression policy, while clarification temporarily adds a fixed
top-$K$ set of retrieved clips.
For longer streams, the searchable history can be limited by a retention policy
such as a FIFO window, which bounds archive size without changing the
TP-KV budget.

\subsection{\texorpdfstring{{Generated-Question Analysis Latency}}{Generated-Question Analysis Latency}}
\label{sec:appendix-clarification-latency}
{}{We measure generated-question latency in the all-query condition.
Question generation is attempted for all 302 Ambiguous queries, of which
212 pass the validity gate and the remaining 90 use Direct Retrieval.} Mean latency is 0.168~s for the Stage-1 check,
0.665~s for question generation,
1.247~s for search-query generation, 0.031~s for archive search,
1.927~s for retrieved-visual prefilling, and 0.910~s for re-answering.
{}{In this all-query condition, total model-side runtime averages 4.948~s,
excluding the evaluation-only GPT-5 gate (0.935~s) and real-user response delays.} Under entropy gating, the Stage-1
check runs for every query, while the remaining clarification steps run only
for selected queries.

\section{\texorpdfstring{Prompt Templates}{Prompt Templates}}
\label{sec:appendix-prompts}

We first list the default prompt structure used in
StreamClarify's core pipeline. The additional prompt, validity gate, and
simulated response used only in the generated-question analysis are listed
separately. Stage~1 produces a binary
\texttt{[Yes]/[No]} pre-answer judgment, and Stage~2 generates the
final answer conditioned on the Stage~1 output. The Multiple-Choice
Constraint is appended to Stage~2 when answer options are provided.
The Search Query Generation prompt is invoked after a clarification is
triggered, converting the clarified user-side guidance into visual search phrases for
{}{archive} retrieval.

\subsection{\texorpdfstring{Core Pipeline}{Core Pipeline}}

\noindent\begin{minipage}[b]{\linewidth}
\noindent\textbf{Stage 1: Pre-Answer Judgment.}
\par\smallskip
{\setlength{\fboxsep}{4pt}
\noindent\raisebox{\depth}{\fcolorbox{black!20}{black!3}{%
\parbox{0.94\linewidth}{\small\ttfamily\raggedright
Output ONLY [Yes] or [No].\par
[Yes] = Your answer will be based on what you SEE in the video
(specific objects, locations, actions).\par
- "Where did the person put the cup?" -> [Yes] if you see the location in video.\par
- "What did the person just do?" -> [Yes] if you observed the action.\par
[No] = Your answer will come from general knowledge (could answer without watching video).\par
- "How does a microwave work?" -> [No].\par
- "What are the benefits of exercise?" -> [No].\par
If your answer references ANY specific detail from the video, use [Yes].
}}}}
\end{minipage}

\medskip
\noindent\begin{minipage}[b]{\linewidth}
\noindent\textbf{Search Query Generation.}
Invoked on the clarification branch after clarified user guidance is available.
\par\smallskip
{\setlength{\fboxsep}{4pt}
\noindent\raisebox{\depth}{\fcolorbox{black!20}{black!3}{%
\parbox{0.94\linewidth}{\small\ttfamily\raggedright
You are uncertain about answering this question based on what you've seen so far.\par
Question: "\{question\}"\par
Generate up to \{max\_phrases\} visual search phrase(s) to find missing but important evidence.\par
Rules:\par
- 3 to 8 words per phrase.\par
- noun-phrase style preferred, with short descriptive verbs allowed (e.g., "open door", "person holding cup").\par
- avoid imperatives/commands (e.g., no "look for", "find", "show").\par
- focus on concrete visible entities/actions/scenes you need to find.\par
- output JSON list of strings only.\par
Output:
}}}}
\end{minipage}

\medskip
\noindent\begin{minipage}[b]{\linewidth}
\noindent\textbf{Stage 2: Answer Generation.}
\par\nopagebreak[4]\smallskip
{\setlength{\fboxsep}{4pt}
\noindent\raisebox{\depth}{\fcolorbox{black!20}{black!3}{%
\parbox{0.94\linewidth}{\small\ttfamily\raggedright
Answer the question directly without any prefixes or bracketed tags.\par
RULES:\par
- If the question asks about THIS VIDEO -> answer from video frames.\par
- If the question asks for GENERAL KNOWLEDGE -> answer using your knowledge directly.\par
Answer with complete sentences. Be accurate and specific.\par
Do not add explanations unless asked.\par
For colors: standard terms only:\par
red/orange/yellow/green/blue/purple\par
pink/brown/black/white/gray/beige.
}}}}
\end{minipage}

\medskip
\noindent\begin{minipage}[b]{\linewidth}
\noindent\textbf{Multiple-Choice Constraint.}
This constraint is appended to Stage~2 when answer options are
provided.
\par\smallskip
{\setlength{\fboxsep}{4pt}
\noindent\raisebox{\depth}{\fcolorbox{black!20}{black!3}{%
\parbox{0.94\linewidth}{\small\ttfamily\raggedright
Multiple-choice response: select one option from the provided choices.\par
Do not generate an answer outside the provided options.\par
Output format: <number>. <option text> (copy exactly). Start with the option number.
}}}}
\end{minipage}

\subsection{\texorpdfstring{{}{Generated-Question Analysis}}{Generated-Question Analysis}}

\medskip
\noindent\begin{minipage}[b]{\linewidth}
\noindent\textbf{Generated-Question Prompt.}
\label{sec:appendix-generated-question-prompt}
{}{This
prompt is used only in the generated-question analysis. It is applied to 162
queries under entropy gating and all 302 queries without gating.}
\par\smallskip
{\setlength{\fboxsep}{4pt}
\noindent\raisebox{\depth}{\fcolorbox{black!20}{black!3}{%
\parbox{0.94\linewidth}{\small\ttfamily\raggedright
You need one piece of additional user-side information before answering.\par
\medskip
Ask exactly one short clarification question that requests only the missing
user intent needed to interpret the latest question.\par
\medskip
Rules:\par
- If a word or phrase has an unclear reference, ask what it refers to.\par
- Do not answer the user's question.\par
- Do not ask the user for information that should be obtained from the video,
such as a location, color, action, state, count, or yes/no fact.\par
- Do not guess, name, or suggest the missing information yourself.\par
- Ask exactly one question.\par
- Use no more than 20 words.\par
- Output only the question.\par
\medskip
Latest user question:\par
\{latest\_user\_question\}\par
\medskip
Clarification question:
}}}}
\end{minipage}

\medskip
\noindent\begin{minipage}[b]{\linewidth}
\noindent\textbf{Evaluation-Only Validity Gate.}
This gate is used only in the controlled generated-question analysis.
It receives the latest user query, the generated question, and the annotated
referent, then returns a binary validity decision. It does not revise the
question or produce a user response. Rejected questions use Direct Retrieval.
\par\smallskip

{\setlength{\fboxsep}{4pt}
\noindent\raisebox{\depth}{\fcolorbox{black!20}{black!3}{%
\parbox{0.94\linewidth}{\small\ttfamily\raggedright
Determine whether the generated question is a valid clarification of the latest
user question and can be answered truthfully using only the provided intended
referent.\par
\medskip
Return valid=true only if it asks for the identity or meaning of the unclear
referent needed to interpret the latest user question.\par
\medskip
Return valid=false if it:\par
- asks for the original QA answer or a video-grounded fact;\par
- asks for information not contained in the intended referent;\par
- targets an unrelated expression;\par
- guesses, names, or suggests a candidate referent;\par
- is a leading yes/no confirmation;\par
- is malformed or is not a clarification question.\par
\medskip
Do not answer either question.\par
Return only the structured binary decision.\par
\medskip
Gate input:\par
\{\par
\ \ "latest\_user\_question": "...",\par
\ \ "generated\_question": "...",\par
\ \ "intended\_referent": "..."\par
\}\par
\medskip
Gate output:\par
\{"valid": true\}
}}}}
\end{minipage}

\medskip
\noindent\begin{minipage}[b]{\linewidth}
\noindent\textbf{Simulated Referent-Only Response.}
For an accepted question, the simulator inserts the annotated referent
into the fixed template below. Generated Clarification and Oracle Guidance
therefore receive the same reply on accepted items.
\par\smallskip
{\setlength{\fboxsep}{4pt}
\noindent\raisebox{\depth}{\fcolorbox{black!20}{black!3}{%
\parbox{0.94\linewidth}{\small\ttfamily\raggedright
I mean \{referent\}.
}}}}
\end{minipage}

\fi

\end{document}